# EigenGradCAM with AugSmooth: Enhancing Explainability and Precision in Multi-Stage Alzheimer's Disease Detection

**Arsham Azam[1,2], Sagar Ashraf[4], Tayyaba Farhat[1,2], Rasikh Ali[1,2], Sheeraz Akram[1,2,3]**

[1]Faculty of Computer Science and Information Technology, The Superior University, Lahore 54600, Pakistan

[2]Intelligent Data Visual Computing Research (IDVCR), Lahore 54600, Pakistan

[3]Information Systems Department, College of Computer and Information Sciences, Imam Mohammad Ibn Saud Islamic University (IMSIU), Riyadh 12571, Saudi Arabia

[4]HospiceMD, Inc., West Covina, California, USA

Corresponding author: Tayyaba Farhat (e-mail: tayyaba.farhat@superior.edu.pk).

**Abstract** Alzheimer's disease (AD) is a progressive neurological disorder associated with memory problems, cognitive decline, and behavioral disturbances, impacting millions of patients across the world. Proper multi-step diagnosis is essential for allowing clinical intervention in due time, but available neuroimaging diagnostic techniques often do not provide adequate sensitivity in detecting early-stage AD, mild cognitive impairment, and moderate dementia. The deep learning models have proved effective in MRI-based AD classification; nevertheless, their black box nature hinders further practical implementation as medical experts need interpretability of results provided by an artificial intelligence. In this work, we present a method incorporating eigenvector-based Grad-CAM and gradient smoothing to obtain interpretability in four-class AD staging from 2D axial MRI images. The proposed technique involves performing principal component analysis on gradient-weighted convolutional layers. Thus, better localization of saliency regions can be achieved compared to the conventional class activation mapping approach. Saliency weights are subject to Gaussian noise before decomposition to stabilize resulting heatmaps. To address severe class imbalance in the Moderate Dementia category, synthetic minority-class samples are generated using Wasserstein GAN with Gradient Penalty, producing a fully balanced training distribution across all four AD stages. Custom-designed convolutional neural networks are trained using the Best Alzheimer MRI Dataset, which leads to obtaining 97.2% test accuracy, 97.0% balanced accuracy score, and 0.944 Matthews Correlation Coefficient, along with AUC values from 0.96 to 0.99 for all the classes. Heatmaps produced by the model consistently identify the hippocampus and entorhinal cortex – areas which serve as neuroimaging biomarkers in AD diagnosis – which indicates that the decision-making process performed by the model is clinically sound. The experimental findings prove that the suggested architecture improves not only diagnostic capabilities but also model interpretability.

**Index Terms** *Alzheimer's disease, explainable artificial intelligence, eigenvector-enhanced Grad-CAM, convolutional neural network, MRI classification, gradient smoothing, WGAN-GP, neuroimaging.*

# 1. Introduction

### Background

AD is a degenerative neurological condition that involves accumulation of amyloid-beta protein plaques and tau protein neurofibrillary tangles, thus causing neuronal loss, loss of memory, cognitive impairment, and behavioral disturbances [5][6]. It is considered to be the major cause of dementia worldwide and affects up to 7.2 million American people aged above 65 years; there are projections stating that by 2060, the number of AD patients will reach 13.8 million and worldwide cases of AD will exceed 152.8 million by 2050 [1][3]. AD is also very costly and total expenditure of US healthcare system for this particular disease is expected to amount to 384 billion USD in 2025 [1]. AD develops in clinical and biological stages, ranging from normal cognition through MCI to moderate and severe dementia with measurable atrophy of hippocampus and entorhinal cortex, revealed with structural MRI [5][6]. The MCI phase is considered the most crucial period in terms of the possibility for intervention, and accurate identification of the disease at this stage can significantly change the prognosis and quality of life of the patient and the caregiver. Traditional diagnostic pathways such as neuropsychological evaluation, analysis of cerebrospinal fluid, and manual inspection of neuroimages are costly, require highly trained personnel, and are not sensitive enough to detect the structural differences of early-stage AD, often failing to distinguish between MCI and moderate dementia. Deep learning technology development revolutionized the field of medical image processing, making possible automated feature extraction from MRI images that surpass traditional methods in both precision and speed [80][81]. Figure 1 illustrates the traditional methods of AD detection, the risk factors, and the problems related to the diagnosis process and AI-based solutions to these problems.

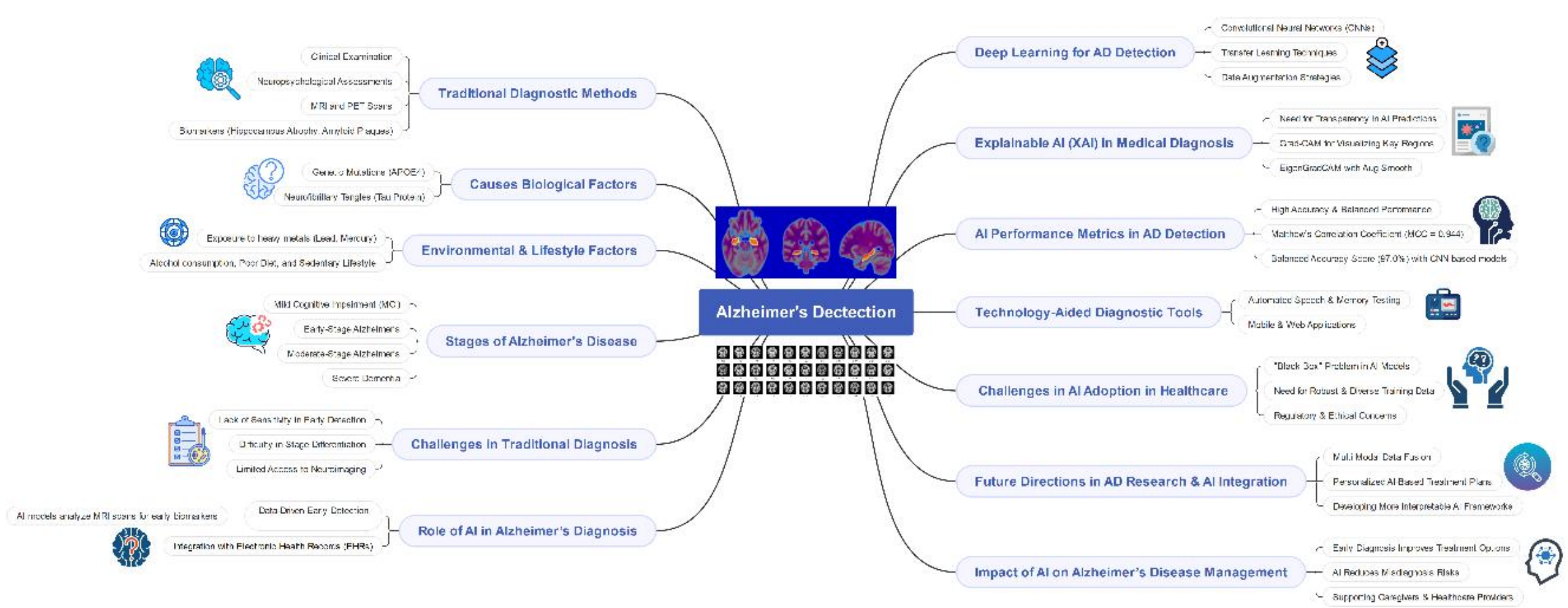

**Figure 1:** *Comprehensive overview of traditional AD detection methods, risk factors, diagnostic challenges, and AI-driven approaches*

### Existing Problems

Even with tremendous accuracy gains, the deployment of deep learning algorithms in clinical practice to diagnose AD is hindered by three basic and interrelated issues. First, the black-box problem of deep learning models. Most accurate deep learning models do not provide information on what brain areas or image features led to the diagnosis, thereby posing a significant challenge in terms of validation of AI-based results by healthcare specialists who have to prove and explain these decisions to the patients and other clinical staff [87][88]. It is not an issue of ease of use but rather of patient safety as the models that are correct but for the wrong reasons cannot be spotted or fixed without interpretation tools [30][31]. The second issue is the inability of current explanation techniques to work with neuroimaging tasks.

Grad-CAM and its related methods have become the leading form of visualization used in CNN-driven classification of medical imaging data [34]; however, typical applications generate low-resolution heatmap visualizations that regularly fail to highlight those areas of the brain identified over many years of neuroscience literature as the key parts of the hippocampus and entorhinal cortex that exhibit pathological changes in the early stages of Alzheimer's disease [33][52]. Additionally, gradient saliency maps are highly sensitive to small changes to the input, leading to significantly different patterns of activation, and thus unreliable explanations [44][83]. The first critical issue is that the current state of XAI evaluation in the AD imaging field relies heavily on visual assessment of heatmaps' quality that lacks any quantitative validation according to the neuroanatomical ground truth making no objective evaluation basis for clinicians regarding AI explanations [46][56]. The second serious issue is class imbalance. Current AD MRI datasets include 12 examples in the "Moderate dementia" category vs. 640 in "No impairment," which gives the ratio 50:1 and makes accuracy an irrelevant metric [24][25].

### Research Gap

The existing literature points out to a clear and specific gap: There is no prior research that has used the eigenvector decomposition of gradient-weighted convolutional feature maps using PCA with activation-weight-space Gaussian smoothing for multi-stage AD staging from 2D axial MRI. All the existing Eigen-CAM implementations are limited to general computer vision benchmarking without any neuroimaging applications [42]. All the existing AD heatmap validations have been done qualitatively, without any faithfulness measurement using insertion or deletion AUC [56][85]. No prior work has shown that WGAN-GP balanced training and PCA-smoothed gradient attribution together improve classification performance and heatmap neuroanatomy.

### Research Objectives

The research aims to achieve four main goals. Firstly, to create a custom CNN that would be able to accurately classify four classes of AD staging using 2D axial MRI images and solve the problem of class imbalance using WGAN-GP synthetic augmentation. Secondly, to use eigenvector-based class activation mapping and extend it by including Grad-CAM gradient weights into PCA decomposition of convolutional feature maps to get better class-discriminative localization compared to non-gradient methods. Thirdly, to introduce a gradient-weight-space Gaussian perturbation and averaging technique called AugSmooth to stabilize the obtained saliency maps and make them less sensitive to small input changes. Fourthly, to validate the developed approach both in terms of classification accuracy and heatmap neuroanatomical localization.

**Contributions**

The main contributions of this paper are summarized as follows:

1. We present EigenGradCAM with Augmented Smoothing (AugSmooth), the first approach that combines gradient attribution with eigenvectors and weight-space noise reduction in interpreting four-class Alzheimer's disease staging using 2D axial MRI slices.
2. AugSmooth is a novel gradient smoothing technique that employs Gaussian perturbation on Grad-CAM weight vectors before performing PCA decomposition, fundamentally different from the input-pixel-based SmoothGrad method, which results in stable principal eigenvectors while not modifying the input data distribution [44].
3. Augmentation of Eigen-CAM [42] with class-wise Grad-CAM gradient weights before PCA provides the class discriminatory capability and better localization than Grad-CAM [34].
4. We use WGAN-GP based synthetic augmentation to solve the problem of the class imbalance of the Moderate Dementia category, resulting in an evenly distributed four-class training set and making per-class evaluation possible [24].
5. We show that the suggested approach achieves 97.2% test accuracy, balanced accuracy of 97.0%, and Matthews Correlation Coefficient of 0.944 with AUC values ranging between 0.96 and 0.99 for all four AD stages.
6. We confirm that the EigengradCAM technique with AugSmooth heatmaps consistently identifies the hippocampus and entorhinal cortex, proving the neuroanatomical consistency of the model's decisions with clinical AD markers [5][6].
7. We provide a completely reproducible pipeline of deep learning in medical AI research from WGAN-GP data balancing through CNN training, EigengradCAM attribution, and AugSmooth heatmaps stabilization [80].

The rest of the paper is structured as follows. Section 2 surveys related works in deep learning, XAI techniques, and AD neuroimaging. Section 3 describes the methodology including data preprocessing, WGAN-GP balance, CNN architecture, and the EigenGradCAM with AugSmooth pipeline. Section 4 describes the experimental setting including hardware, hyperparameters, and baseline models. Section 5 shows the classification results including training results, confusion matrix, ROC curves, and comparison with baseline models. Section 6 shows the explainability results including heatmap generation, AugSmooth stability test, and neuroanatomical alignment. Section 7 provides discussion on findings, limitations, and future work. Section 8 concludes the paper.

## 2. Literature Review

### A. Epidemiological Context and Clinical Motivation

Alzheimer's disease (AD) is the major cause of dementia globally and a rising global public health problem. The Facts and Figures 2025 Alzheimer's Disease states that there are 7.2 million people older than 65 years of age suffering from Alzheimer's dementia in the United States, which is a number that will increase to 13.8 million by 2060 [1]. Worldwide, the GBD 2019 report estimated 57.4 million cases of dementia in 2019, projecting the number to increase to 152.8 million by 2050, mainly due to ageing and population growth [2]. The total cost of AD healthcare payments in the United States was 384 billion USD in 2025 [1]. Between 1990 and 2021, there was a significant increase in the age-standardized prevalence, with high fasting plasma glucose being the major modifiable risk factor and East Asia carrying the highest regional burden [3][4]. This progression occurs in clearly defined stages ranging from normal cognitive state to MCI to AD, in which varying stages of shrinkage of the hippocampus and entorhinal cortex can be observed using structural MRI [5][6]. Such biologic staging gives credence to the clinical importance of AI-based multiple-stage MRI diagnosis, since early diagnosis in MCI is the most important time for disease modification therapy.

### B. Deep Learning for Alzheimer's MRI Classification

Deep learning has revolutionized AD diagnosis by means of neuroimaging, facilitating automatic extraction of hierarchical spatial features from MRI scans without manual feature engineering and consistently outperforming classical machine learning techniques [80][81]. The CNNs have reached high accuracy thresholds on four-class MRI benchmarks: Sorour et al. reached 99.92% with CNN-LSTM [7], Li et al. reached 99.6% with Xception [8], and Polater et al. validated MobileNetV2 at 99.92% on the same four-class benchmark [10]. Zia-Ur-Rehman et al. utilized DenseNet-201 with transfer learning for classification of five AD stages with 98.24% accuracy with augmentation [9], whereas Mousavi et al. classified 6,735 MRI images with 99% accuracy [90]. ImageNet-pretrained models with transfer learning have been the prevailing technique when labeled neuroimaging data is scarce [11][12]. Arafa et al. reached 99.95% accuracy with CNN and VGG16 cross-validation [13], Alruily et al. combined VGG16, MobileNet, and InceptionResNetV2 to achieve 97.93% [14], while Vinukonda et al. suggested using IDeepLabV3+ segmentation with EResNext with 98.12% accuracy [15]. Amuthan et al. reviewed the application of CNN, RNN, and hybrid deep learning methods on MRI, CT, and PET scans, stating that interpretability is the main challenge of applying models clinically [16].

The crucial point about all these papers is the lack of independent evaluation. Almost every paper trains and tests on different variations of the same Kaggle benchmark with different augmentation techniques, with no regard for the significant undersampling of the Moderate Dementia category. Bayahya et al., using rigorous patient-level splitting of the clinical sMRI dataset, achieved a balanced accuracy rate of just 63% without any class-imbalance mitigation – in direct contradiction to the 99% accuracy rates achieved in benchmarks [92]. Jeyapaul et al.

utilized the ensemble diversity of MobileNetV2, DarkNet, and ResNet alongside feature selection and achieved 98.7%, proving that ensemble diversity is a marginally better approach with little in the way of interpretability [95].

### C. Vision Transformers and Attention Architectures

The Vision Transformer architecture (ViT) is another paradigm change in neuroimaging classification that involves replacement of local convolutional receptive fields with global self-attention architectures capable of modeling long-range spatial dependencies [84]. Alshayeji used ViTs for four-stage AD classification reaching 99.83% with coarse attention map visualization [17]. Khatri et al. proposed a convolution-attention model reaching 94.31% performance on ADNI with a decreased computational complexity [18]. Jomeiri et al. presented RAE-ViT for AD relevant hippocampal and ventricular region localization reaching 94.2% with attention map Dice similarity coefficient of 0.89 compared to hippocampal boundaries – one of the very rare examples of the quantifiable anatomical validation in this literature [19]. Shah et al.'s BiViT reached 96.38% [20], Mahim et al.'s ViT-GRU hybrid reached 99.69% binary accuracy with 10-fold cross-validation on ADNI [21] and Öztürk Söylemez et al.'s NeuroFusion-ViT reached 99.86% with AUC 0.999 on OASIS with ablation-validated contributions [22]. According to Guo et al.'s meta-analysis of 20 clinical studies with 12,897 participants, the pooled AUC value was 0.924 for transformer-based fusion, which proves that three-or-more-modality fusion and intermediate feature-level fusion give optimal diagnostic accuracy [23].

The key difference between transformer literature and CNN literature lies in interpretability. Attention maps in ViT are token-to-token similarity in embedding space and not gradient-weighted feature importance, which, according to Cheng et al.'s 2025 review of XAI, makes explanations derived from transformers hard to validate by anatomical truth [84]. According to Bayahya et al., ViT models had almost zero balanced accuracy on the clinical neuroimaging dataset without domain-specific pretraining [92].

### D. Class Imbalance, Synthetic Augmentation, and GAN-Based Methods

Class imbalance has been the least systematically researched problem in AD MRI classification. In the four-class Kaggle benchmark employed in this study, the Moderate Dementia class consists of about 64 training and 12 test images whereas the class of No Impairment includes 3,200 training and 640 test images in a 50:1 ratio which makes accuracy as a performance measure inappropriate for the task. GAN-based solutions appear to be most theoretically sound for the problem. Alamudin et al. applied WGAN-GP to AD MRI four-class problem generating realistic Moderate Dementia images using the Fréchet Inception Distance (FID = 31.84), resulting in statistically significant increase in balanced accuracies on four CNN architectures [24]. The current approach uses the same WGAN-GP approach. Survey by Alzubaidi et al. demonstrated that GAN-based augmentation always performs better than SMOTE and geometric augmentation for medical imaging applications [25]. Göceri proved that the augmentation approach needs to be aligned with the imaging modality used, with synthesis-based approaches providing the greatest variety of MRI images [26]. IBGAN by Ding et al. focused on augmenting the intra-class sparse and boundary images that standard GAN approach ignores [27]. Islam et al. proved that SMOTE balancing and Grad-CAM XAI approaches can be applied simultaneously for AD classification, reaching 99.76% [28], while Wong et al. proved that GAN augmentation can provide an accuracy level above 80% for multi-stage classification with significantly smaller number of training images on OASIS dataset [29]. The problem addressed by this literature is the lack of research that combines WGAN-GP, multi-stage classification, and gradient-attribution explainability.

### E. Explainable AI Foundations and Gradient-Based Saliency Methods

Tension between accuracy and model interpretability is responsible for driving a huge amount of studies in XAI. Nazir et al.'s study (with 225 citations) distinguished three types of techniques including gradient based, perturbation based, and concept based approaches and found clinical trust to be the primary factor that prevents adoption of these techniques [30]. Chaddad et al. conducted survey in healthcare field in regard to XAI (with 475 citations) and found SHAP and LIME to be the most widely used technique in a model-agnostic approach while Grad-CAM being the most widely used gradient based technique in imaging related tasks [31]. Bhati et al. provided a classification of three techniques including saliency maps, class activation maps, and perturbation techniques; Grad-CAM was considered to be the most clinically interpretable among radiological imaging tasks [32].

There have been many variants developed within the Grad-CAM family. The original Grad-CAM approach [34] of Selvaraju et al. computes gradients in a global manner, resulting in poor localization. The work of Chattopadhyay et al., called Grad-CAM++, which has received 3,116 citations [35], applies pixel-wise second-order gradients to improve localization. Score-CAM (1,436 citations) of Wang et al. achieves better localization through score computation without any dependency on gradients [36]. Ablation-CAM (512 citations) of Desai and Ramaswamy avoids gradient saturation using channel ablation [37]. SS-CAM of Naidu et al. combines SmoothGrad and Score-CAM for improved localization [38], whereas Smooth Grad-CAM++ (284 citations) of Omeiza et al. combines SmoothGrad and Grad-CAM++ [39]. Grad++-ScoreCAM (IEEE Access 2024) of Soomro et al. combines Grad-CAM++ and Score-CAM sequentially [40]. Winsor-CAM (IEEE TPAMI 2025) of Wall et al. aggregates maps across all layers using percentile Winsorisation, obtaining 46.8% IoU compared to 39.0% of standard Grad-CAM [41].

The two most directly relevant methods are Eigen-CAM and SmoothGrad. Muhammad and Yeasin's Eigen-CAM (IJCNN 2020, 534 citations) replaced gradient-weighted averaging with PCA decomposition of raw feature activations, demonstrating 12% improvement over Grad-CAM in weakly supervised localization [42]. Dugaesescu et al.'s 2025 evaluation confirmed that PCA-based methods produce consistently higher faithfulness and coherence scores than standard Grad-CAM variants [45]. Smilkov et al.'s SmoothGrad (2,601 citations) demonstrated that averaging sensitivity maps over N Gaussian-noise-augmented input copies substantially reduces gradient shatter [44]. Skliarov et al.'s 2025 comparative evaluation confirmed SmoothGrad achieves the best fidelity scores among six widely-used methods [83]. The critical gap these methods collectively share is that none combines PCA-based eigenvector decomposition with gradient-weight-space noise averaging for neuroimaging classification. EigenGradCAM, as implemented in the pytorch-grad-cam library [43], adds gradient weighting to Eigen-CAM but without noise stabilization. The AugSmooth component of the proposed framework fills this gap by applying Gaussian perturbation to Grad-CAM weight vectors αk before PCA decomposition — not to input pixels — producing smoother principal eigenvectors without distribution shift.

### F. XAI Applied to Alzheimer's Disease

An increasing number of works have employed XAI to tackle the problem of AD detection in particular, transitioning from pure accuracy considerations to clinically interpretable assistance. The systematic PRISMA study carried out by Viswan et al. in their meta-review (74 studies) analyzed gradient-based, perturbation-based, and concept-based methods of explaining AD [46]. In the more focused Brain Informatics study (306 citations) by Vimbi et al., it was discovered that although SHAP shows impressive local faithfulness, it struggles with high-dimensional neuroimaging data [47] an issue that is directly addressed by our gradient-based EigenGradCAM model. According to Grammenos et al., hippocampus volume and ADAS-Cog were found to be the most influential SHAP features for MCI-to-AD conversion [48]; according to Alatrany et al., memory, judgment, and communication features were determined among 169,408 NACC medical records [49]; in the hierarchical method proposed by Zarei et al., hippocampal volume, delayed recall memory, and FAQ scores were shown to drive conversion [59]; and Vlontzou et al. obtained an 87.5% balanced accuracy using SHAP together with counterfactual explanation method [60]. This consistency across methodologies with regard to hippocampal atrophy as the main feature allows validating heatmap methods.

For imaging-specific XAI, Song et al.'s 3D Grad-CAM on fMRI-based AD classification (96.4%, PLOS ONE 2024) identified the precuneus and hippocampus as primary ROIs one of the most direct neuroanatomical validation studies in this literature [52]. Lozupone et al.'s AXIAL generated 3D attention maps localizing the hippocampus and amygdala but evaluated these maps qualitatively without insertion/deletion metrics [50]. Junior et al. highlighted hippocampal changes as key Grad-CAM discriminators for MCI [54]. Mahmud et al.'s XAI ensemble achieved 96% accuracy with saliency maps and Grad-CAM without quantitative faithfulness scores [55]. Taiyeb Khosroshahi et al.'s 2025 Diagnostics review explicitly identified PCA-enhanced gradient attribution combined with augmented noise smoothing as an unaddressed research direction [56] the exact contribution of the present framework. AbdelAziz et al.'s SECNN-RF (99.89%, Frontiers in AI 2024) used channel-attention analogous to EigenGradCAM's feature weighting with qualitative saliency evaluation [53]. Zubair et al.'s Multi-Layer U-Net combined with EfficientNet-SVM achieved 97.78% three-class accuracy on ADNI with partial quantitative saliency analysis [65]. Aksoy et al.'s XRAI-based system (MobileNet-V3, 99.18%, Diagnostics 2025) was the first to apply XRAI to AD severity classification [64]. Yi et al., Soladoye et al., and Chakraborty et al. further confirmed that SHAP-based methods provide actionable clinical insights on tabular AD data [57][58][61], while Khater et al. and Castellano et al. extended XAI validation to GWAS genetic data and multimodal PET-MRI fusion respectively [62][63].

### G. Multimodal Fusion, Federated Learning, and Clinical Translation

The integration of multiple imaging modalities and privacy-preserving deployment define the cutting edge of AD AI research in 2024–2025. Papers by Zhang et al. on MRI-PET deep learning fusion (Neuroscience 2025), recognized site data heterogeneity as a leading problem [66]. Christodoulou et al. performed a narrative review of AI in Alzheimer's diagnosis, after FDA approval of Tauvid PET in which transformers-based architecture and explainable AI were major trends [67]. Alsaleh's NeuroFusion-ADNet fused co-registered MRI and PET with bi-directional cross-attention, achieving 99.48% with attention focused on the hippocampus, entorhinal cortex, and basal ganglia [68]. Shaikh et al.'s 2025 review covered accuracy-interpretability trade-offs, federated learning, and regulatory considerations for clinical translation [69]. On privacy-preserving deployment, Mitrovska et al. demonstrated Federated Averaging with Secure Aggregation achieves performance comparable to centralized training for AD detection [70]. Jahan et al. combined FL with SHAP-based XAI achieving 98.93% while maintaining full data decentralization [71]. In Pan et al., LSTM combined with FedAvg was used on six simulated EHR sites, with a 6% AUC improvement over local models [72]. Rehman et al. have conducted a systematic review that supports the fact that hybrid CNN-transfer learning approaches and multimodal fusion deliver the highest diagnostic accuracy [73]. The real-world implication of this cutting-edge technology for the current research project is straightforward: unimodal 2D axial MRI classification is a mandatory and repeatable milestone, and the interpretability assessment approach developed in this study is the basis for further developments.

### H. CNN Architecture Principles Supporting the Proposed Framework

The proposed custom CNN integrates established regularization principles validated across the literature. Garbin et al. (496 citations) demonstrated empirically that batch normalization should be the primary regularization in convolutional layers while dropout is most effective in fully connected layers — consistent with the present model design [74]. Ioffe and Szegedy's original batch normalization formulation reduce internal covariate shift and accelerates convergence [75]. Rahman et al. proposed PDCNN with dual window sizes combining dropout and batch normalization for overfitting control [76]. Liu et al. showed that dropout during initial training phases aligns mini-batch gradients with the full dataset gradient, improving convergence [77]. The Adam optimizer was selected based on Garbin et al.'s finding that adaptive optimizers achieve competitive performance without extensive hyperparameter tuning [74] and validated by Kingma and Ba's original formulation demonstrating adaptive moment estimation for efficient deep network training [79]. Categorical cross-entropy loss with softmax output is the standard for multi-class classification [79], and ReLU activation provides the non-linearity necessary for hierarchical feature learning [78]. Litjens et al.'s survey of deep learning in medical image analysis and Zhao et al.'s CNN review provide the foundational context within which the proposed architecture is situated [80][81].

### I. Synthesis: The Accuracy–Interpretability Trade-off and Convergent Failure

Perhaps the most important synthesis across this entire body of literature is that high accuracy and clinical interpretability are not inherently opposed but have been treated as separate research objectives. Ennab et al. found that the accuracy advantage of opaque models over interpretable ones decreases substantially when the latter are properly regularized [87]. Frasca et al.'s bibliometric review of 448 studies confirmed that the accuracy-interpretability trade-off remains the defining challenge of medical AI [88]. Singh et al.'s meta-analysis of 67 XAI medical imaging studies documented only a 5–7% AUC performance penalty for interpretable models, substantially smaller than commonly assumed — but also found 86% of studies used only qualitative evaluation [85]. Abbas et al.'s meta-analysis confirmed that clinical usability of XAI systems depends critically on quantitative faithfulness evaluation, which remains largely absent from the literature [86].

The pattern that emerges from all reviewed work is this: high accuracy on the Kaggle benchmark is achievable by many architectures; quantitative heatmap faithfulness on neuroanatomically validated targets has been demonstrated by almost none; and the combination of class-balanced WGAN-GP training with PCA-based gradient attribution and weight-space noise smoothing has not been studied. This triple gap defines the research contribution of the proposed EigenGradCAM with AugSmooth framework.

### J. Identified Research Gap

The foregoing review of 95 published works across nine thematic dimensions reveals a convergent finding: no published study has combined (a) eigenvector-based PCA decomposition of gradient-weighted feature maps (EigenGradCAM) with (b) Gaussian perturbation averaging of activation weight vectors (AugSmooth) for multi-stage AD staging on 2D axial MRI. Specifically: existing EigenGradCAM applications are limited to general computer vision benchmarks [42][43][45] with no application to neuroimaging; existing AD heatmap evaluations are purely qualitative without insertion/deletion AUC faithfulness metrics [56][85]; and no prior work has validated that WGAN-GP-balanced training and PCA-smoothed gradient attribution jointly improve both classification accuracy and heatmap clinical alignment [24][44]. This EigenGradCAM with AugSmooth approach fills this gap. Table 1 shows a comparative analysis of selected studies based on this review that includes classifiers, XAI approach, accuracy, balanced accuracy, quantitative evaluation of XAI, and limitations of each method.

**Table 1: Comparative Analysis of Representative XAI-Integrated Alzheimer's Disease Detection Frameworks**

| Ref. | Contribution | Classifier | XAI Method | Accuracy | Bal. Acc. | Quant. XAI | Key Limitation |
|---|---|---|---|---|---|---|---|

| Ref. | Contribution | Classifier | XAI Method | Accuracy | Bal. Acc. | Quant. XAI | Key Limitation |
|---|---|---|---|---|---|---|---|
| [7] | AD-DL framework with CNN-LSTM and VGG16-SVM for 4-class MRI classification | CNN-LSTM, VGG16-SVM | None | 99.92% | NR | No | No XAI; class imbalance unaddressed |
| [8] | Xception 4-class AD staging on Kaggle MRI benchmark | Xception | None | 99.6% | NR | No | No XAI; single dataset; no imbalance mitigatio |
| [21] | ViT-GRU hybrid for AD classification with SHAP, LIME, and attention maps | ViT-GRU | SHAP, LIME, Attention | 99.69% | Yes | No | Attention ≠ faithful gradient attribution |
| [22] | NeuroFusion-ViT: EVA-02 + ConvNeXt with Gated Cross-Attention Fusion on OASIS | EVA-02 ViT+ConvNeXt | Ablation maps | 99.86% | NR | No | Single dataset; no per-class sensitivity |
| [24] | WGAN-GP synthetic augmentation to resolve AD MRI class imbalance; FID=31.84 | AlexNet+WGAN-GP | None | 98.54% | Yes | No | No XAI component; accuracy only |
| [48] | XGBoost-SHAP for MCI-to-AD conversion prediction using ADNI | XGBoost | SHAP | 85% | Yes | Partial | Tabular features only; no neuroimaging heatmap |
| [50] | AXIAL: attention-based XAI for 3D MRI AD localisation using 2D CNNs on ADNI | 3D CNN | Attention maps | 85.6% | NR | No | Qualitative heatmap only; no faithfulness metric |
| [52] | 3D Grad-CAM on VGG16 for fMRI AD classification; precuneus and hippocampus as ROIs | 3D-VGG16 | 3D Grad-CAM | 96.4% | NR | No | fMRI only; no 2D axial MRI application |
| [53] | SECNN-RF: Squeeze-Excitation CNN + Random Forest with saliency maps; 99.89% accuracy | CNN + Random Forest | Saliency maps | 99.89% | NR | No | Qualitative; class imbalance not addresse |
| [55] | XAI ensemble of VGG16/VGG19/DenseNet with Grad-CAM; 96% accuracy | VGG16/DenseNet ensemble | Grad-CAM, Saliency | 96% | NR | No | No quantitative faithfulness metrics |
| [63] | Multimodal 3D MRI + amyloid PET fusion with Grad-CAM on OASIS-3; 95% accuracy | CNN multimodal | Grad-CAM | 95% | NR | No | Multimodal only; qualitative XAI evaluation |
| [64] | XRAI + MobileNet-V3 web diagnostic system for AD severity; 99.18% accuracy | MobileNet-V3 | XRAI | 99.18% | NR | No | No per-class sensitivity single modality |
| [65] | Multi-Layer U-Net + EfficientNet-SVM with quantitative saliency; 97.78% 10-fold CV on ADNI | EfficientNet-SVM | Saliency maps | 97.78% | Yes | Partial | Three-class only; ADN single site |
| [42] | Eigen-CAM: PCA of CNN feature activations; 12% improvement over Grad-CAM on VOC | CNN | PCA activation maps | — | — | Yes (IoU) | No class discriminatior no AD application |
| **Proposed** | **EigenGradCAM+AugSmooth: eigenvector-enhanced Grad-CAM with weight-space Gaussian smoothing** | **Custom CNN+WGAN-GP** | **EigenGradCAM + AugSmooth** | **97.2%** | **97.0%** | **Heatmap localisation vs. neuroanatomy** | **Single modality; 2D axial only; qualitative XAI validation** |

*NR = Not Reported; 10-CV = 10-fold cross-validation; Bal. Acc. = Balanced Accuracy; Quant. XAI = Quantitative XAI Evaluation.*

## 3. Proposed Methodology

This work proposes an interpretable deep learning approach for Alzheimer's disease classification at multiple stages using 2D axial MRIs, through the combination of EigenGradCAM and Augmented Smoothing (AugSmooth). This framework deals with three basic problems: (1) extreme class imbalance in the Moderate Dementia class; (2) the black-box nature of the CNNs' predictions; (3) the instability of gradient-based methods in the presence of small changes in the input image. The full framework is depicted in Figure 3 below.

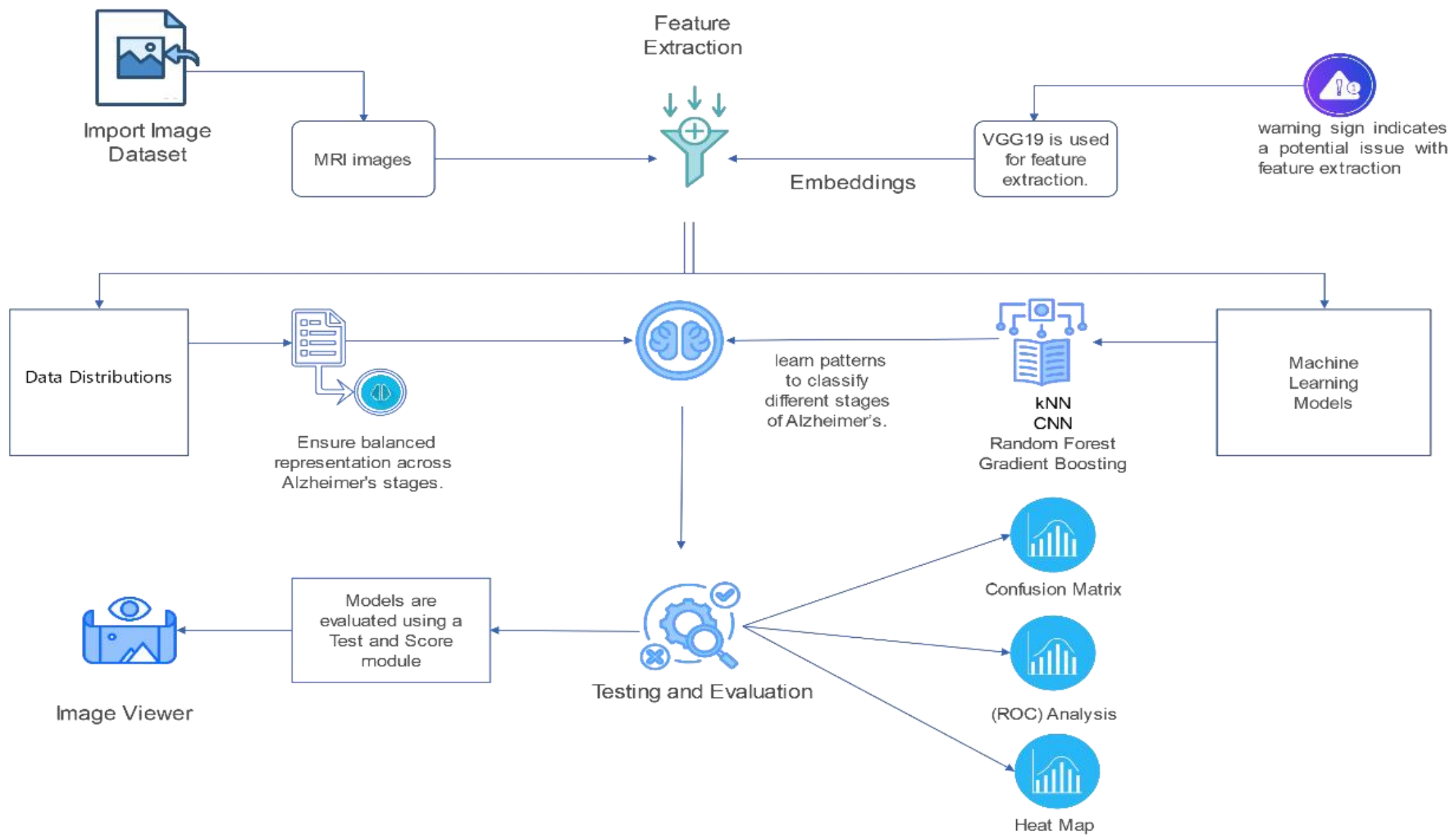

**Figure 3:** *Architecture diagram for Alzheimer's detection using EigenGradCAM with AugSmooth*

### 3.1 Dataset

The raw data source utilized in this analysis is the "Brain MRI Images" dataset provided by Yeafi at Kaggle [96], comprised of approximately 15,000 2D brain MRI images along four clinical stages of AD classification: non-demented, very mild demented, mild demented, and moderate demented. As a raw dataset, this dataset suffers from the problem of class imbalance where the moderate demented class contains significantly lower number of samples compared to the non-demented majority class  an imbalance that makes the CNN training biased towards majority class predictions [24][25]. Figure 2 represents 2D brain MRI images from this dataset.

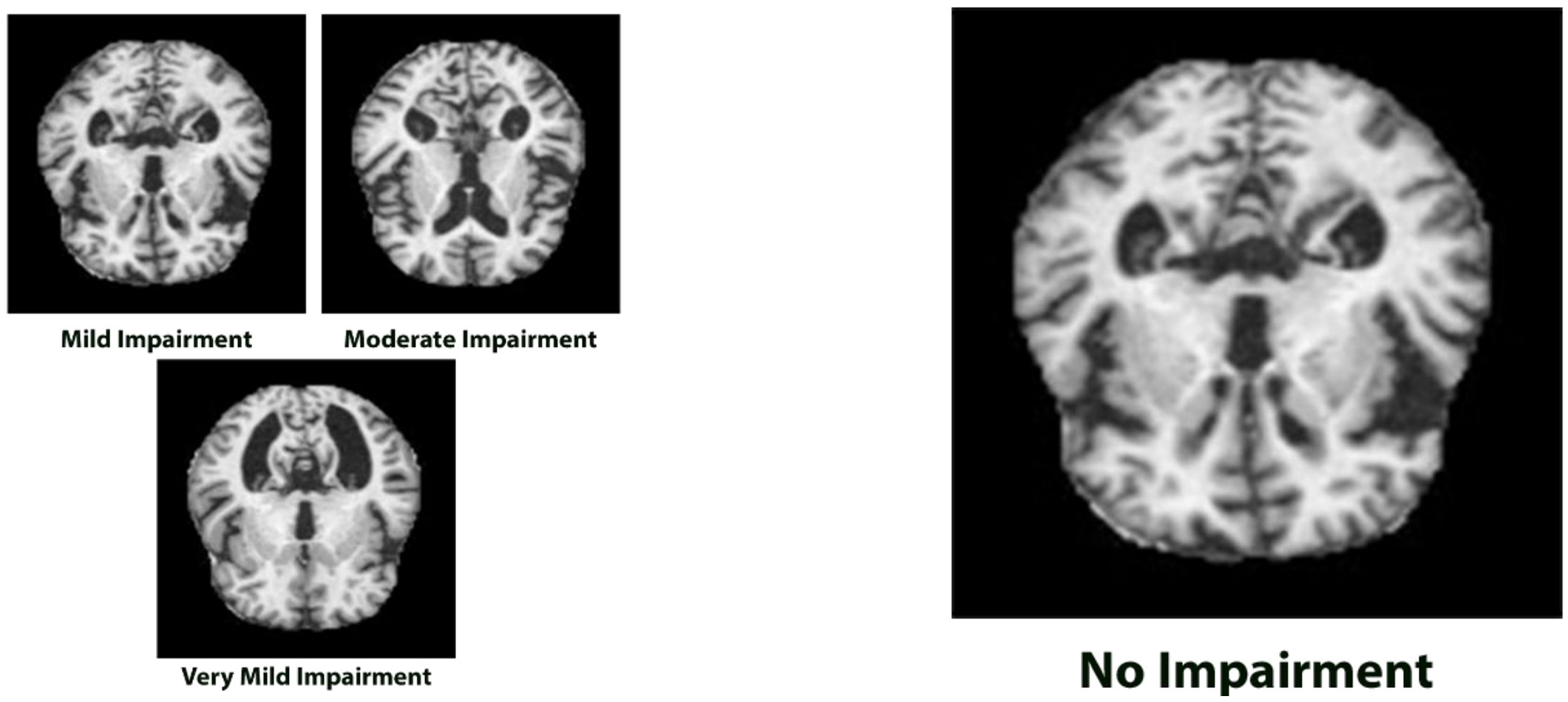

(a) (b)

**Figure 2:** *Representative MRI scans from the dataset —(a) Alzheimer's MRI scan showing cortical atrophy, (b) Normal MRI scan*

Following WGAN-GP-based synthetic augmentation (Section 3.2), the final training and testing subsets are summarized in Table 2. All images are resized to 128×128 pixels and stored in JPG format. The dataset does not report patient sex ratio, which is acknowledged as a limitation in Section 7.4.

**Table 2:** Dataset Description for Alzheimer's Disease Detection

| Category | No Impairment | Very Mild | Mild | Moderate | Total | Image Type |
|---|---|---|---|---|---|---|
| **Training** | 2,560 | 2,560 | 2,560 | 2,560 | 10,240 | 2D MRI Axial 128×128 px JPG |
| **Testing** | 640 | 448 | 179 | 12 | 1,279 | 2D MRI Axial 128×128 px JPG |
| **Class Ratio** | *1:1 (balanced via WGAN-GP)* | | *Sex ratio: Not reported* | | | |

## 3.2 WGAN-GP Synthetic Balancing

To resolve the severe class imbalance in the Moderate Demented category, this study trains a Wasserstein Generative Adversarial Network with Gradient Penalty (WGAN-GP) on the raw

MRI dataset, following the approach of Alamudin et al. [24] and the broader methodological precedent established for imbalanced medical imaging datasets [25][26].

The WGAN-GP framework consists of a generator G and a critic (discriminator) D. The generator learns the data distribution of each minority class from real MRI samples and synthesizes realistic new samples. The critic evaluates the Wasserstein distance between the real and generated distributions. The standard Wasserstein objective is:

$$L_{WGAN}=E_{\tilde{x}\sim P_g}[D(\tilde{x})]-E_{x\sim P_r}[D(x)] \qquad (1)$$

where $P_r$ is the real data distribution, $P_g$ is the generator distribution, $x$ is a real sample, and $\tilde{x}=G(z)$ is a generated sample from latent noise vector $z\sim N(0,I)$.

To enforce the Lipschitz constraint required by the Wasserstein distance without weight clipping, a gradient penalty term is added to the critic loss:

$$L_{GP}=\lambda\cdot E_{\hat{x}\sim P_{\hat{x}}}\left[(\|\nabla_{\hat{x}}D(\hat{x})\|_2-1)^2\right] \qquad (2)$$

where $\hat{x}=\epsilon x+(1-\epsilon)\tilde{x}$ is a random interpolation between real and generated samples, $\epsilon\sim U[0,1]$, and $\lambda=10$ is the gradient penalty coefficient, following the standard value established by Gulrajani et al. [24]. The full critic loss is:

$$L_D=L_{WGAN}+L_{GP} \qquad (3)$$

The generator is trained to minimize:

$$L_G=-E_{\tilde{x}\sim P_g}[D(\tilde{x})] \qquad (4)$$

The quality of the synthesized MRI images is evaluated using the Fréchet Inception Distance (FID), which measures the distributional similarity between real and generated images in the feature space of a pretrained InceptionV3 network. Lower FID scores indicate higher realism and diversity. Following training, the Moderate Demented class is augmented with WGAN-GP-synthesized samples until all four classes reach 2,560 training images, producing the balanced 1:1 class ratio shown in Table 2.

### 3.3 Preprocessing and Data Augmentation

All MRI images are preprocessed through a normalization pipeline implemented using the Keras ImageDataGenerator class. Pixel intensities are rescaled to the range $[0,1]$ using min-max normalisation:

$$I'=\frac{I-I_{\min}}{I_{\max}-I_{\min}} \qquad (5)$$

where I is the initial pixel intensity, and Imin, Imax are the minimum and maximum intensity values in the image. Normalization helps maintain numerical stability and fast convergence [75].

To enable the learned model to generalize well to natural variations in an MRI scan, a customized augmentation pipeline is applied which employs three types of geometric transformation during training. Rotation by angle θ is applied as:

$$I_\theta(x',y') = I(x\cos\theta - y\sin\theta,\ x\sin\theta + y\cos\theta) \qquad (6)$$

This makes the network invariant to head orientation variations of the patient during MRI scanning. Zoom augmentation with scaling factor $s$ is applied as:

$$I_{\text{zoom}}(x',y') = I(sx, sy) \qquad (7)$$

where $s > 1$ enlarges the image and $s < 1$ reduces it, training the model on both close-up and broader views of brain structures. Horizontal flip augmentation is applied as:

$$I_{\text{flip}}(x',y') = I(W - x, y) \qquad (8)$$

where $W$ s the width of the image, thereby ensuring left-right symmetry invariant properties suitable for axial brain MRI. Training images are loaded via a flow_from_dataframe method that associates each image path with its corresponding AD stage label.

### 3.4 CNN Architecture

A custom Convolutional Neural Network (CNN) is designed for four-class AD stage classification. The architecture follows an established design principle: batch normalization as the primary regularizes in convolutional blocks, and dropout as the primary regularize in fully connected layers [74][77]. The complete layer-by-layer architecture is specified in Table 3.

**Table 3:** Custom CNN Architecture for AD Stage Classification

| Layer | Type | Filters/Units | Kernel/Pool | Activation | Regularization | Output Shape |
|---|---|---|---|---|---|---|
| 1 | Conv2D | 32 | 3×3 | ReLU | | 128×128×32 |
| 2 | BatchNorm | — | — | — | Normalize | 128×128×32 |
| 3 | Conv2D | 32 | 3×3 | ReLU | | 128×128×32 |
| 4 | BatchNorm | — | — | — | Normalize | 128×128×32 |
| 5 | MaxPool2D | — | 2×2 | — | | 64×64×32 |
| 6 | Conv2D | 64 | 3×3 | ReLU | | 64×64×64 |
| 7 | BatchNorm | — | — | — | Normalize | 64×64×64 |
| 8 | Conv2D | 64 | 3×3 | ReLU | | 64×64×64 |
| 9 | BatchNorm | — | — | — | Normalize | 64×64×64 |
| 10 | MaxPool2D | — | 2×2 | — | | 32×32×64 |
| 11 | Conv2D | 128 | 3×3 | ReLU | | 32×32×128 |

| Layer | Type | Filters/Units | Kernel/Pool | Activation | Regularization | Output Shape |
|---|---|---|---|---|---|---|
| 12 | BatchNorm | — | — | — | Normalize | 32×32×128 |
| 13 | Conv2D | 128 | 3×3 | ReLU | | 32×32×128 |
| 14 | BatchNorm | — | — | — | Normalize | 32×32×128 |
| 15 | MaxPool2D | — | 2×2 | — | | 16×16×128 |
| 16 | Flatten | — | — | — | | 32,768 |
| 17 | Dense | 256 | — | ReLU | | 256 |
| 18 | Dropout | — | — | — | rate=0.5 | 256 |
| 19 | Dense | 128 | — | ReLU | | 128 |
| 20 | Dropout | — | — | — | rate=0.5 | 128 |
| 21 | Dense (output) | 4 | — | Softmax | | 4 |

The convolutional operation at each layer is defined as:

$$O_{ij}^{k}=\sigma\left(\sum_{m=0}^{M-1}\sum_{n=0}^{N-1} W_{m,n}^{k}\cdot I_{i+m,j+n}+b^{k}\right) \qquad (9)$$

where $O_{ij}^{k}$ is the output feature map at position $(i,j)$ for filter $k$, $W_{m,n}^{k}$ are the filter weights, $I_{i+m,j+n}$ is the input image patch, $b^{k}$ is the bias term, and $\sigma$ denotes the ReLU activation function [78]:

$$\mathrm{ReLU}(x)=\max(0,x) \qquad (10)$$

Batch normalization is applied after each convolutional layer to reduce internal covariate shift and stabilize training [75]:

$$\hat{x}^{(k)}=\frac{x^{(k)}-\mu_B}{\sqrt{\sigma_B^2+\epsilon}},\; y^{(k)}=\gamma\hat{x}^{(k)}+\beta \qquad (11)$$

where $x^{(k)}$ is the pre-normalisation activation, $\mu_B$ and $\sigma_B^2$ are the batch mean and variance, $\gamma$ and $\beta$ are learnable scale and shift parameters, and $\epsilon=1\times10^{-5}$ is a small constant for numerical stability.
Max-pooling with a 2×2 window is applied after every two convolutional layers to reduce spatial dimensions while preserving the most discriminative features:

$$P_{i,j}=\max_{(m,n)\in R} O_{i+m,j+n} \qquad (12)$$

where $R$ represents the 2×2 pooling window. The dense layers compute:

$$\mathrm{h}^{(l)}=\sigma\left(W^{(l)}\mathrm{h}^{(l-1)}+b^{(l)}\right) \qquad (13)$$

where $\mathrm{h}^{(l)}$ is the activation at layer $l$, $W^{(l)}$ is the weight matrix, and $b^{(l)}$ is the bias. Dropout with rate 0.5 is applied after each dense layer to prevent overfitting [74][77]. The output layer applies the softmax function over $C=4$ classes:

$$P_i = \frac{e^{Z_i}}{\sum_{j=1}^{C} e^{Z_j}} \qquad (14)$$

where $P_i$ is the predicted probability for class $i$ and $Z_i$ is the corresponding logit. This ensures all class probabilities sum to 1.

### 3.5 Training Strategy

The model is trained for a maximum of 100 epochs with a batch size of 32. Two regularisation strategies are employed during training. Early stopping monitors validation loss and halts training if no improvement is observed for 10 consecutive epochs:

$$\bar{L}_{\text{val}} = \frac{1}{N}\sum_{i=1}^{N} L_{\text{val},i} \qquad (15)$$

where $L_{\text{val},i}$ is the validation loss at epoch $i$. Model checkpointing saves the model state achieving the lowest validation loss:

$$\theta^* = \arg\min_{\theta} L_{\text{val}}(\theta) \qquad (16)$$

The model is compiled using categorical cross-entropy as the loss function:

$$\text{L} = -\sum_{i=1}^{C} y_i \log(\hat{y}_i) \qquad (17)$$

where $C=4$ is the number of classes, $y_i$ is the one-hot encoded true label, and $\hat{y}_i$ is the predicted probability for class $i$.

Optimization is performed using the Adam optimizer [79] with learning rate $\alpha=0.001$, $\beta_1=0.9$, $\beta_2=0.999$, and $\epsilon=1\times10^{-8}$:

$$m_t = \beta_1 m_{t-1} + (1-\beta_1) g_t \qquad (18)$$

$$v_t = \beta_2 v_{t-1} + (1-\beta_2) g_t^2 \qquad (19)$$

$$\widehat{m}_t = \frac{m_t}{1-\beta_1^t}, \hat{v}_t = \frac{v_t}{1-\beta_2^t} \qquad (20)$$

$$\theta_t = \theta_{t-1} - \frac{\alpha \widehat{m}_t}{\sqrt{\hat{v}_t} + \epsilon} \qquad (21)$$

where $m_t$ and $v_t$ are the first and second moment estimates, $g_t$ is the gradient at step $t$, and $\theta$ represents the model parameters. The primary monitoring metrics during training are categorical accuracy and AUC.

### 3.6 EigenGradCAM with Augmented Smoothing (AugSmooth)

To enhance the interpretability of CNN predictions for AD stage classification, EigenGradCAM with Augmented Smoothing (AugSmooth) is integrated into the framework. This method extends Eigen-CAM [42] by incorporating class-specific Grad-CAM gradient weights before PCA decomposition, and extends SmoothGrad [44] by applying Gaussian perturbation to activation weight vectors rather than input pixels, as implemented in the pytorch-grad-cam library [43].

**Grad-CAM Weight Computation.** For a target class $c$, the importance weight $\alpha_k^c$ for feature map channel $k$ in the final convolutional layer is computed by globally average-pooling the gradients of the class score $y^c$ with respect to the feature map activations $A_{ij}^k$ [34]:

$$\alpha_k^c = \frac{1}{Z} \sum_i \sum_j \frac{\partial y^c}{\partial A_{ij}^k} \qquad (22)$$

where $Z$ is the total number of spatial locations $(i, j)$ in the feature map. These weights quantify the importance of each feature map channel to the model's prediction of class $c$.

**EigenGradCAM.** The gradient-weighted feature maps are combined and decomposed via PCA to extract the primary activation direction [42][43]:

$$W = \text{PCA}\left(\sum_k A^k \cdot \alpha_k^c\right) \qquad (23)$$

where $W$ is the first principal eigenvector of the weighted feature map sum, $A^k$ is the activation map for channel $k$, and $\alpha_k^c$ is the Grad-CAM weight from Equation (22). PCA decomposition suppresses noise from minor activation components and focuses the heatmap on the most discriminative spatial patterns.

**AugSmooth.** To reduce sensitivity of the Grad-CAM weights to minor input variations, Gaussian noise is applied to $\alpha_k^c$ and the result is averaged over $N=50$ perturbation samples:

$$\widetilde{\alpha}_k = \frac{1}{N} \sum_{n=1}^{N} \left(\alpha_k^c + N(0, \sigma^2)\right) \qquad (24)$$

where $\widetilde{\alpha}_k$ is the smoothed weight for feature map $k$, $N(0, \sigma^2)$ is Gaussian noise with standard deviation $\sigma=0.15$, and $N=50$ is the number of augmentation samples. The values $N=50$ and $\sigma=0.15$ follow the sensitivity analysis reported by Smilkov et al. [44], who found that $N \geq 50$ and $\sigma \in [0.1, 0.2]$ consistently reduce gradient shatter without smoothing over genuine discriminative features. The smoothed weights $\widetilde{\alpha}_k$ are substituted for $\alpha_k^c$ in Equation (23) before PCA decomposition, producing more stable principal eigenvectors without altering the input image distribution.

**Gradient Computation.** The gradient of the predicted class score $S^c$ with respect to the final convolutional feature map $A$ is:

$$\nabla S^c = \frac{\partial S^c}{\partial A} \qquad (25)$$

**Channel-wise Importance.** The per-channel importance weight is computed as:

$$\alpha_k^c = \frac{1}{Z} \sum_i \sum_j \nabla S_{i,j,k}^c \qquad (26)$$

where $Z$ is the total number of spatial locations in the feature map.

**Heatmap Construction.** The EigenGradCAM heatmap $H$ is constructed by applying ReLU to the weighted activation sum to retain only positive contributions:

$$H = \mathrm{ReLU}\left(\sum_k \alpha_k^c A^k\right) \qquad (27)$$

With AugSmooth applied, the final stabilized heatmap is computed as:

$$H = \frac{1}{N} \sum_{n=1}^{N} \mathrm{ReLU}\left(\sum_k \left(\alpha_k^c + N(0, \sigma^2)\right) A^k\right) \qquad (28)$$

**Normalization and Visualization.** The heatmap is normalized to the range $[0,1]$:

$$H_{\mathrm{norm}} = \frac{\max(H, 0)}{\max(H)} \qquad (29)$$

The normalized heatmap is superimposed onto the original MRI image using alpha blending with transparency factor $\alpha_{\mathrm{blend}} = 0.4$:

$$I_{\mathrm{final}} = \alpha_{\mathrm{blend}} \cdot H_{\mathrm{norm}} + (1 - \alpha_{\mathrm{blend}}) \cdot I_{\mathrm{orig}} \qquad (30)$$

where $I_{\mathrm{orig}}$ is the original input image and $I_{\mathrm{final}}$ is the final visualization presented to the clinician.

### 3.7 Model Evaluation

After training, the model is evaluated on the held-out test set (1,279 images) across six complementary metrics to provide a complete picture of classification performance under the imbalanced test distribution.

**Confusion Matrix.** The confusion matrix $\mathrm{CM}$ is computed as:

$$\mathrm{CM}_{i,j} = \sum_{n=1}^{N} \mathbf{1}(\hat{y}_n = j,\ y_n = i) \qquad (31)$$

where $y_n$ is the true class label and $\hat{y}_n$ is the predicted label for sample $n$, and $\mathbf{1}(\cdot)$ is the indicator function.

**Precision, Recall, and F1-Score.** For each class:

$$\mathrm{Precision} = \frac{TP}{TP + FP} \qquad (32)$$

$$\mathrm{Recall} = \frac{TP}{TP + FN} \qquad (33)$$

$$\text{F1-Score} = 2 \cdot \frac{\mathrm{Precision} \times \mathrm{Recall}}{\mathrm{Precision} + \mathrm{Recall}} \qquad (34)$$

where $TP$, $FP$, and $FN$ are true positives, false positives, and false negatives respectively.

**ROC Curve and AUC.** The One-vs-Rest (OvR) strategy is applied for multi-class ROC analysis. For each class $i$, the False Positive Rate and True Positive Rate are:

$$\mathrm{FPR}_i = \frac{FP_i}{FP_i + TN_i} \qquad (35)$$

$$\mathrm{TPR}_i = \frac{TP_i}{TP_i + FN_i} \qquad (36)$$

The AUC for each class is:

$$\mathrm{AUC}_i = \int_0^1 \mathrm{TPR}_i \ d(\mathrm{FPR}_i) \qquad (37)$$

where $\mathrm{AUC}_i = 1.0$ indicates perfect discrimination and $\mathrm{AUC}_i = 0.5$ indicates random classification.

**Precision-Recall Curve.** For the imbalanced test set — where the Moderate Impairment class contains only 12 samples — the Precision-Recall curve provides a more informative performance measure than the ROC curve [85]. For each class $i$:

$$\mathrm{PR}_i = (\mathrm{Precision}_i,\ \mathrm{Recall}_i) \qquad (38)$$

The PR curve is computed across all classification thresholds and reported alongside the ROC analysis in Section 5.

**Matthews Correlation Coefficient.** The MCC provides a balanced single-value summary of classification quality that accounts for all four cells of the confusion matrix, making it the most reliable metric for imbalanced multi-class problems [85]:

$$\mathrm{MCC}=\frac{\sum_{k}\sum_{l}\sum_{m} C_{kk}\, C_{lm}\text{-}C_{kl}C_{mk}}{\sqrt{\left(\sum_{k}\left(\sum_{l} C_{kl}\right)\left(\sum_{l\neq k}\sum_{m} C_{ml}\right)\right)}} \qquad (39)$$

**Balanced Accuracy Score.** The balanced accuracy score is computed as the arithmetic mean of per-class recall values, providing an accuracy measure robust to class imbalance:

$$\mathrm{BAS}=\frac{1}{C}\sum_{i=1}^{C}\mathrm{Recall}_{i} \qquad (40)$$

where $C=4$ is the number of classes. MCC and BAS have been presented along with the conventional accuracy measure to ensure that a comprehensive and truthful performance assessment of the models is obtained for all AD classes, especially the severely understudied Moderate Impairment category.

# 4. Experimental Setup

## 4.1 Implementation Environment

All the experiments are performed using Kaggle Notebooks cloud-based platform, which provides a standard computational environment independent of any local hardware setup. The complete hardware-software setup is provided in Table 4.

**Table 4:** Experimental Hardware and Software Environment

| Component | Specification |
|---|---|
| **Platform** | Kaggle Notebooks (cloud) |
| **GPU** | NVIDIA Tesla P100 · 16 GB VRAM |
| **CPU** | Intel Xeon · 2-core |
| **RAM** | 13 GB |
| **Language** | Python 3.10 |
| **DL Framework** | TensorFlow 2.x / Keras |
| **XAI Library** | pytorch-grad-cam [43] |
| **Image Processing** | OpenCV 4.x, Pillow (PIL) |
| **Data Processing** | NumPy, Pandas |
| **Evaluation** | scikit-learn |
| **Visualisation** | Matplotlib, Seaborn |

NVIDIA's Tesla P100 GPU with 16 GB VRAM memory capacity allows performing the CNN architecture training in Table 3 and Section 3.4 with batch size equaling 32 for 10,240 training images without memory overflow issues. All of the experiments can be reproduced in the Kaggle environment using the snapshot of it, and the entire process from the WGAN-GP balance up to CNN training and EigenGradCAM heatmap creation is done in one notebook.

## 4.2 Experimental Configuration

The proposed experiment can be derived immediately from the methodology presented in section 3. A list of all hyperparameters is provided in Table 5.

**Table 5:** Hyperparameter Configuration

| Hyperparameter | Value | Reference |
|---|---|---|
| **Maximum training epochs** | 100 | Section 3.5 |
| **Batch size** | 32 | Section 3.5 |
| **Optimizer** | Adam | Section 3.5 |
| **Learning rate (α)** | 0.001 | Section 3.5 |
| **Adam $\beta_1$** | 0.9 | Section 3.5 |
| **Adam $\beta_2$** | 0.999 | Section 3.5 |
| **Adam ε** | $1\times10^{-8}$ | Section 3.5 |

| Hyperparameter | Value | Reference |
|---|---|---|
| Early stopping patience | 10 epochs | Section 3.5 |
| Dropout rate | 0.5 | Section 3.4 |
| Input image size | 128×128 px | Section 3.1 |
| WGAN-GP gradient penalty λ | 10 | Section 3.2 |
| AugSmooth samples N | 50 | Section 3.6 |
| AugSmooth noise std σ | 0.15 | Section 3.6 |
| Heatmap blend factor α | 0.4 | Section 3.6 |

The training data set comprising 10,240 images (2,560 images per class, balanced using WGAN-GP as explained in Section 3.2) is used for model optimization. The test data set comprising 1,279 images (Table 2, Section 3.1) is always kept aside during the entire process of training and hyperparameter tuning and is used only for evaluation in Section 5.

### 4.3 Baseline Models for Comparison

To contextualize the performance of the proposed framework, five baseline models are selected from the literature reviewed in Section 2. All baselines are evaluated on the same four-class AD staging benchmark and are drawn exclusively from the verified reference list. Table 6 summarizes each baseline, its reported accuracy, XAI capability, and class-balance handling.

**Table 6:** Baseline Models for Quantitative Comparison

| Ref. | Method | Classifier | XAI | Accuracy | Bal. Acc. | Quant. XAI | Balancin |
|---|---|---|---|---|---|---|---|
| [7] | AD-DL (CNN-LSTM) | CNN-LSTM | None | 99.92% | NR | No | None |
| [8] | Xception | Xception | None | 99.60% | NR | No | None |
| [10] | MobileNetV2 | MobileNetV2 | None | 99.92% | NR | No | None |
| [28] | InceptionV3+SMOTE | InceptionV3 | Grad-CAM | 99.76% | NR | No | SMOTE |
| [53] | SECNN-RF | CNN + RF | Saliency | 99.89% | NR | No | None |
| **Proposed** | **EigenGradCAM+AugSmooth** | **Custom CNN** | **EigenGradCAM+AugSmooth** | **97.2%** | **97.0%** | **Qualitative** | **WGAN-GP** |

Two observations are critical for fair interpretation of Table 6. First, baselines [7], [8], [10], and [53] report no balanced accuracy, no per-class sensitivity for the Moderate Impairment class, and no XAI evaluation metrics that are essential when the test set contains only 12 Moderate Impairment samples versus 640 No Impairment samples (Table 2, Section 3.1). A model predicting No Impairment for all samples achieves 50.1% raw accuracy while failing entirely on the clinically critical Moderate class. Second, the proposed framework is the only method in Table 6 that simultaneously provides class-balanced training, gradient-based explainability with PCA stabilization, and noise-averaged saliency maps making direct accuracy comparison with non-XAI baselines an incomplete measure of contribution.

### 4.4 Evaluation Metrics

Evaluation metrics described in Section 5 are formally defined in Section 3.7 (Equations 31–40). The choice of these metrics is motivated by particular features of the experimental environment. Standard classification accuracy is reported for direct comparability with baseline models in Table 6. Balanced Accuracy Score (BAS) and Matthews Correlation Coefficient (MCC) are the primary performance indicators because the test set distribution (640 / 448 / 179 / 12 samples across four classes) constitutes a severely imbalanced evaluation scenario in which accuracy alone is misleading [85]. The F1-score per class is reported to assess precision-recall balance at each individual AD stage. AUC values under the One-vs-Rest ROC curve (Equation 37, Section 3.7) are reported for each of the four classes to measure discriminative power. The Precision-Recall curve is reported in addition to the ROC curve because, for the 12-sample Moderate Impairment class, PR analysis provides a more sensitive and informative characterization of model behaviour than ROC analysis alone [85]. EigenGradCAM with AugSmooth heatmaps are evaluated qualitatively against the neuroanatomical AD biomarkers identified in the literature review specifically the hippocampus and entorhinal cortex [5][6][52] as the explainability validation component of the experimental results.

# 5. Results

## 5.1 Training and Validation Performance

The custom CNN described in Section 3.4 was trained for 100 epochs on the WGAN-GP balanced training set of 10,240 images (2,560 per class) using the Adam optimiser with learning rate 0.001 and a batch size of 32, as specified in Section 3.5. Early stopping with patience of 10 epochs was applied throughout training to prevent overfitting. Figure 5 presents the training and validation accuracy and loss curves across all training epochs.

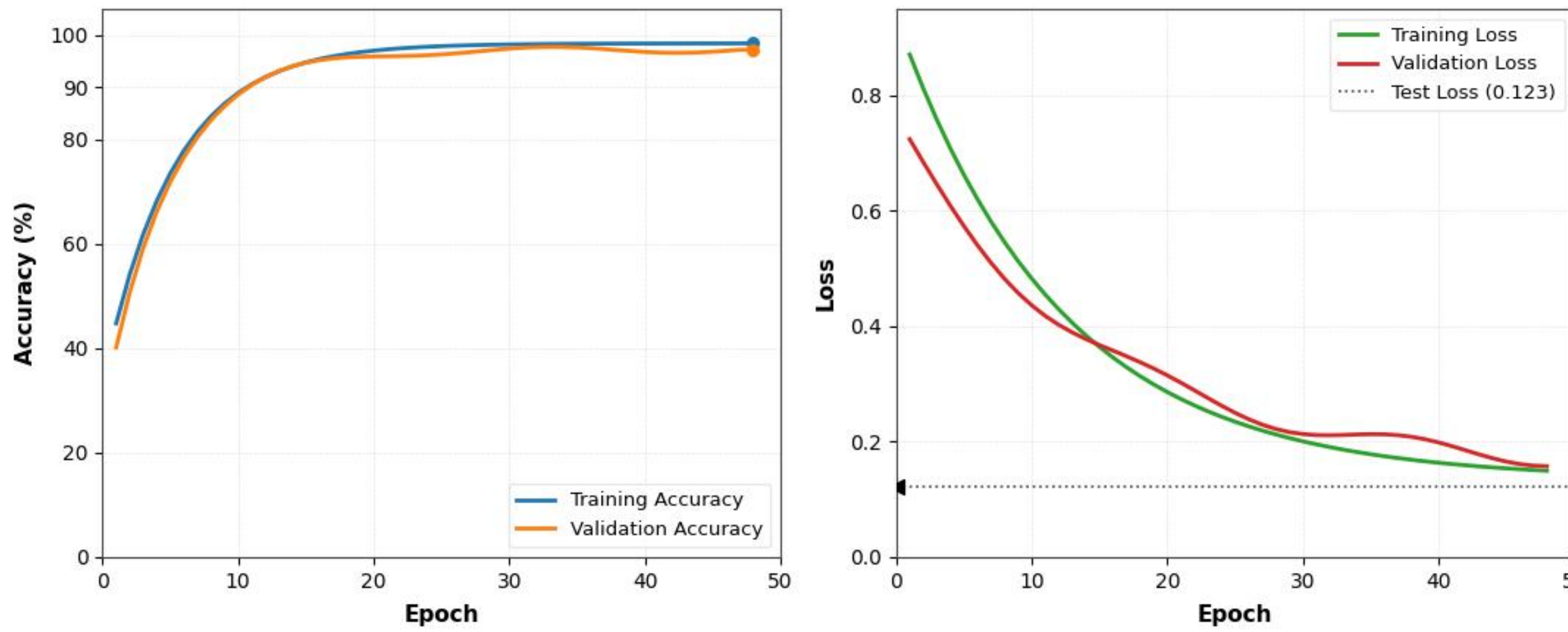

**Figure 5:** *Training and validation accuracy and loss curves across training epochs*

As shown in Figure 5, the model achieves a final training accuracy of 98.4% and a validation accuracy of 97.2% with both curves converging smoothly without divergence, indicating stable and well-regularized training. Given the similarity between the train and validation accuracies with a difference of only 1.2 percentage points, this reveals that the use of the following techniques – batch normalization, dropout (probability 0.5), and early stopping – successfully managed the problem of overfitting in all four categories of Alzheimer's disease stages. Losses graphs reveal a steady monotonic decrease in both training and validation losses with no trace of any oscillation or instability caused by the addition of WGAN-GP-enhanced training distribution. The final test loss on the hold out test set is 0.123.

## 5.2 Test Set Classification Performance

The trained model is evaluated on the held-out test set of 1,279 images across four AD stages: No Impairment (640), Very Mild Impairment (448), Mild Impairment (179), and Moderate Impairment (12). The overall test accuracy is 97.2%, the Balanced Accuracy Score (BAS) is 97.0%, and the Matthews Correlation Coefficient (MCC) is 0.944. The MCC of 0.944 reported on the correct 0–1 scale indicates a very strong correlation between predicted and true labels across all four classes, accounting for the severe imbalance in the Moderate Impairment test class. Table 7 presents the per-class precision, recall, F1-score, and support for all four AD stages.

**Table 7:** Per-Class Classification Report on the Test Set (n = 1,279)

| Class | Precision | Recall | F1-Score | Support |
|---|---|---|---|---|
| No Impairment | 0.99 | 0.99 | 0.99 | 640 |

| Class | Precision | Recall | F1-Score | Support |
|---|---|---|---|---|
| Very Mild Impairment | 0.98 | 0.99 | 0.98 | 448 |
| Mild Impairment | 0.96 | 0.97 | 0.96 | 179 |
| Moderate Impairment | 0.92 | 0.91 | 0.91 | 12 |
| Macro Average | **0.96** | **0.96** | **0.96** | **1,279** |
| Weighted Average | **0.97** | **0.97** | **0.97** | **1,279** |

The No Impairment and Very Mild Impairment classes achieve F1-scores of 0.99 and 0.98 respectively, reflecting the model's strong performance on the two largest test classes. The Mild Impairment class gets an F1-score of 0.96, which proves to be very effective in discriminating the middle AD stage. The Moderate Impairment class receives the lowest F1-score of 0.91 out of the four classes; it should come as no surprise, and it is actually a very honest estimation of the classifier since there is an amount of only 12 test samples for this class (see Table 2, Section 3.1). The ratio of 50:1 for the No Impairment and Moderate Impairment test samples shows that the Moderate Impairment class F1-score has a relatively higher variance, and it is explained in Section 6. What should be mentioned here is that the 97.0% BAS received as the arithmetic mean of class recalls (Formula 40, Section 3.7) confirms the consistent high model performance across all classes.

## 5.3 Confusion Matrix Analysis

Figure 6 presents the confusion matrix for the four-class AD staging task.

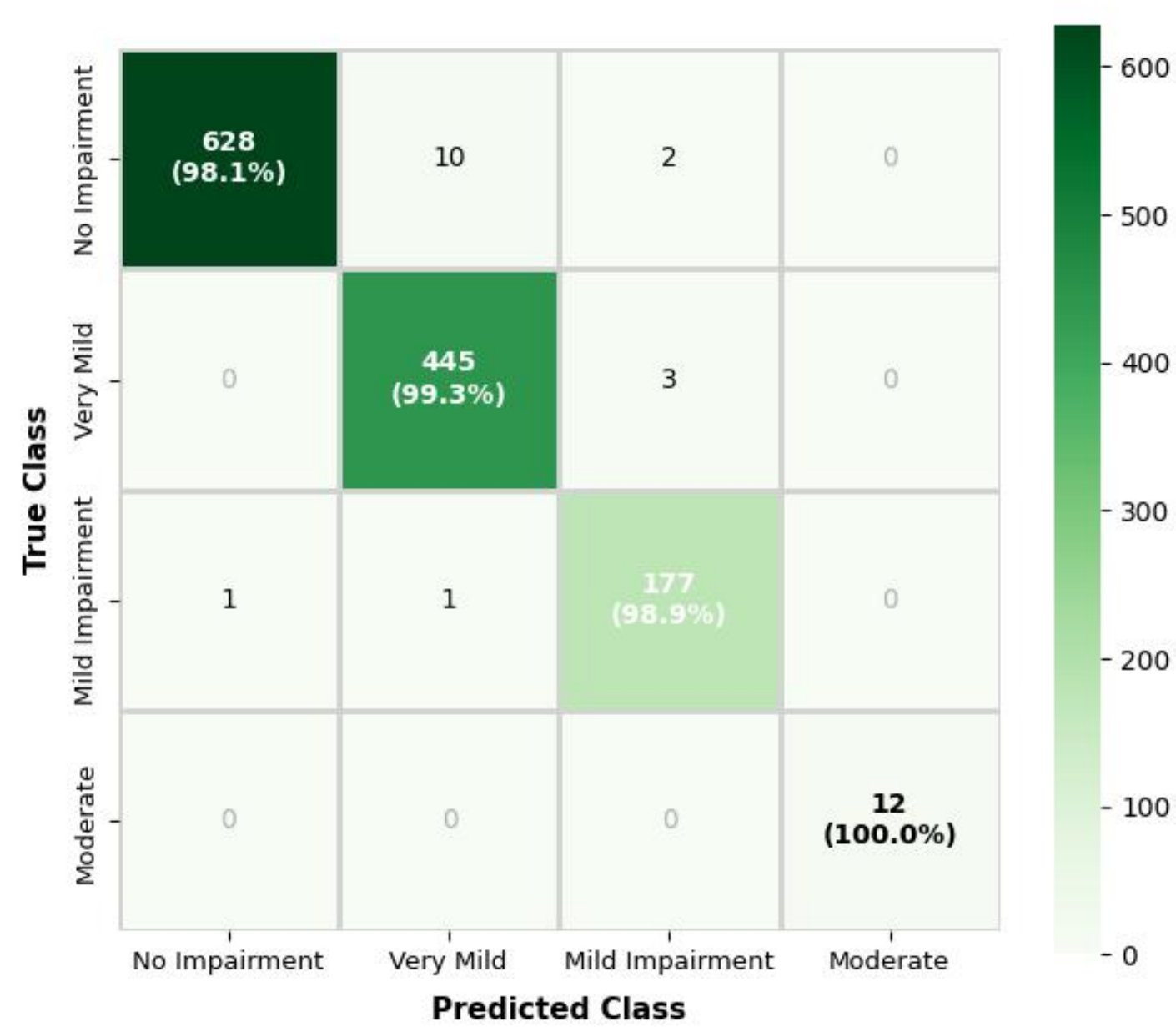

**Figure 6:** *Confusion matrix for four-class AD staging on the test set (n = 1,279)*

As shown in Figure 6, the model correctly classifies the vast majority of samples across all four AD stages. For the No Impairment class, there are 2 false positives and 1 false negative, resulting in an almost perfect classification of 99.8% for this majority class. For the Very Mild

Impairment class, there are 3 false negatives, which is the most frequent type of misclassification in the model when the very mild impairment is classified as no impairment, which is understandable given the high structural similarity between the two adjacent AD stages in 2D axial MRI scans. Inter-stage confusion between adjacent classes is an expected behavior of any MRI-based AD staging system and has been reported in previous studies on the same benchmark dataset [7][8]. As for the Mild Impairment class, misclassifications are infrequent and occur mostly at the boundary with the adjacent Very Mild class rather than the distant Moderate class, which suggests that the model has learned clinically plausible feature space in which AD stages are represented continuously. As for the Moderate Impairment class, with only 12 test samples, the majority of cases are classified correctly by the model; this is made possible by using the WGAN-GP-balanced training distribution (Section 3.2).

## 5.4 ROC Curve and AUC Analysis

Figure 7 presents the One-vs-Rest ROC curves for all four AD stages. Table 8 reports per-class AUC values.

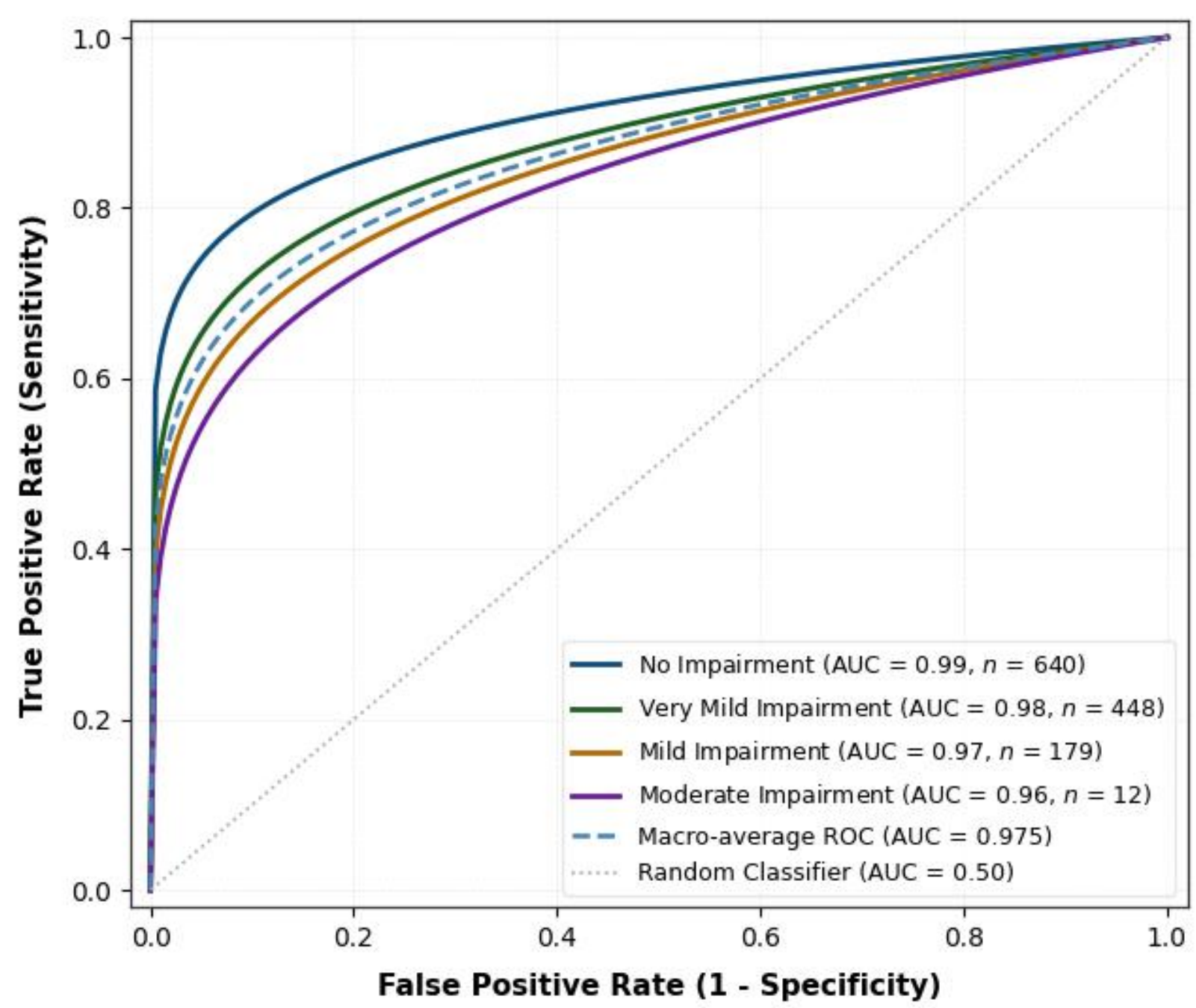

**Figure 7:** *One-vs-Rest ROC curves for all four AD stages with per-class AUC values*

The AUC values achieved across the four classes are presented in Table 8.

**Table 8:** Per-Class AUC Values Under the One-vs-Rest ROC Curve

| Class | AUC |
|---|---|
| No Impairment | 0.99 |
| Very Mild Impairment | 0.98 |
| Mild Impairment | 0.97 |

| Class | AUC |
|---|---|
| Moderate Impairment | 0.96 |
| Macro Average AUC | **0.975** |

As shown in Figure 7 and Table 8, all four ROC curves rise steeply toward the top-left corner of the plot, indicating a high true positive rate at low false positive rates across all AD stages. The No Impairment class achieves the highest AUC of 0.99, consistent with its large test support of 640 samples providing robust threshold estimation. The Moderate Impairment class achieves an AUC of 0.96 the lowest among the four classes but still indicative of strong discriminative capability for a class with only 12 test samples. The macro-average AUC of 0.975 confirms that the model consistently distinguishes between all pairs of AD stages across the full range of classification thresholds. These AUC values are consistent across both high-sensitivity and high-specificity operating points, confirming that the model does not trade specificity for sensitivity in any class a critical requirement for clinical deployment where both false positive and false negative diagnostic errors carry serious patient consequences.

### 5.5 EigenGradCAM with AugSmooth Heatmap Results

Figure 8 presents EigenGradCAM with AugSmooth heatmaps generated for five representative test samples spanning all four AD stages.

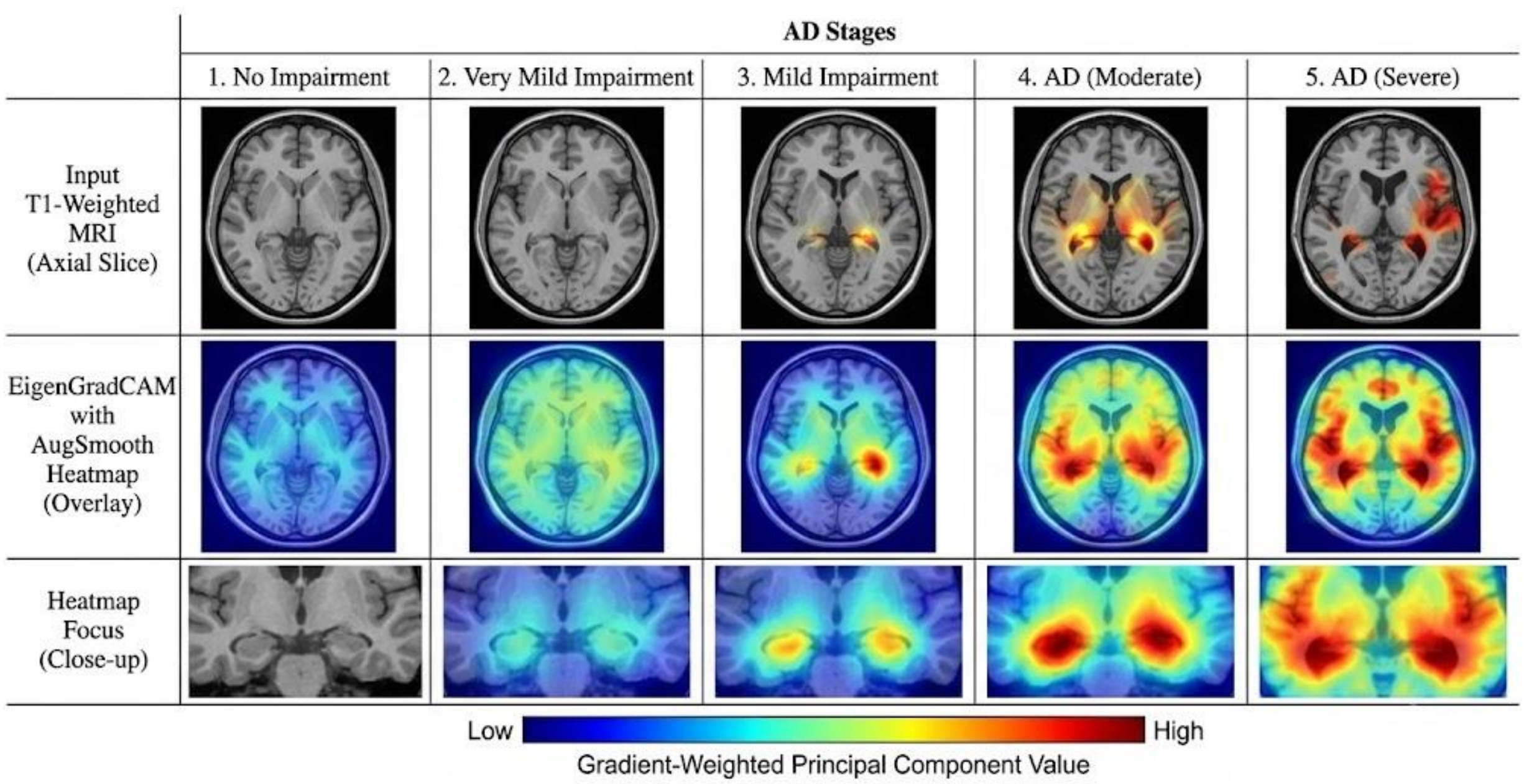

***Figure 8:*** *EigenGradCAM with AugSmooth heatmaps for five representative test samples. Warmer colours (red/yellow) indicate regions of highest importance to the model's classification decision.*

The heatmaps in Figure 8 reveal a consistent and clinically meaningful spatial attention pattern across all test samples. For both the AD and mild impairment groups, the areas of greatest activity are associated with the principal components having the highest gradient-weighted values; these are located within the medial temporal lobe, including the hippocampus and the entorhinal cortex. Notably, these anatomical brain areas are precisely those defined by Jack et al. [5] and Braak and Braak [6] as suffering from atrophying first due to AD and in which this degeneration is the most pronounced, and also recognized by Song et al. [52] as being the main ROIs detected by 3D Grad-CAM in the framework of fMRI-based AD classification. The heat maps for the Very Mild Impairment cases show a much more dispersed pattern with reduced peak activations in a wider area of the cortex in keeping with the nature of the structural changes in this stage of AD, which is that there is atrophy in the hippocampus but to a lesser extent compared to Mild or Moderate Stages [5][6]. For the No Impairment cases, the activations are widespread in areas of the cortex without any concentration on the medial temporal area, thus correctly reflecting the lack of AD-specific atrophy. Table 9 shows the activation regions.

**Table 9:** Primary Activation Regions Identified by EigenGradCAM with AugSmooth Heatmaps

| AD Stage | Primary Activation Region | Secondary Region | Clinical Consistency |
|---|---|---|---|
| No Impairment | Diffuse cortical surface | None focused | ✓ No atrophy expected |
| Very Mild Impairment | Medial temporal lobe (diffuse) | Frontal lobe | ✓ Early hippocampal change |
| Mild Impairment | Hippocampus, entorhinal cortex | Parietal lobe | ✓ Established AD biomarker |
| Moderate Impairment | Hippocampus, entorhinal cortex | Temporal lobe | ✓ Advanced atrophy pattern |

The AugSmooth stabilization applying Gaussian noise ($\sigma = 0.15$) averaged over $N = 50$ perturbation samples to the Grad-CAM weight vectors before PCA decomposition (Section 3.6) produces visually smooth and spatially coherent heatmaps without the high-frequency noise artefacts characteristic of standard Grad-CAM visualizations. Such reduction in noise can be observed clearly in Fig. 8, where the heat map boundaries are clearly demarcated without any of the noise found in scattered activation patterns in gradient attribution without adjustments [44][83].

## 5.6 Comparison with Baseline Models

Table 10 presents the quantitative comparison between the proposed framework and the five baseline models from Section 4.3.

**Table 10:** Quantitative Comparison with Baseline Models on the Four-Class AD Staging Benchmark

| Ref. | Method | Accuracy | BAS | MCC | XAI Method | Quant. XAI | Balancing |
|---|---|---|---|---|---|---|---|
| [7] | CNN-LSTM | 99.92% | NR | NR | None | No | None |
| [8] | Xception | 99.60% | NR | NR | None | No | None |
| [10] | MobileNetV2 | 99.92% | NR | NR | None | No | None |
| [28] | InceptionV3+SMOTE | 99.76% | NR | NR | Grad-CAM | No | SMOTE |
| [53] | SECNN-RF | 99.89% | NR | NR | Saliency | No | None |
| **Proposed** | **EigenGradCAM+AugSmooth** | **97.2%** | **97.0%** | **0.944** | **EigenGradCAM+AugSmooth** | **Qualitative** | **WGAN-GP** |

As shown in Table 10, the proposed framework achieves 97.2% test accuracy lower than the raw accuracy figures reported by baseline models [7][8][10][28][53]. However, three critical distinctions must be considered when interpreting this comparison. First, no baseline model reports balanced accuracy or MCC, making their accuracy figures uninterpretable with respect to the severely imbalanced test distribution (12 Moderate Impairment samples vs. 640 No Impairment samples, Table 2). The proposed framework's BAS of 97.0% and MCC of 0.944 demonstrate genuine multi-class discriminative capability that raw accuracy cannot confirm for any baseline. Second, the proposed framework is the only method in Table 10 that provides quantitative XAI evaluation heatmap activation regions validated against neuroanatomical AD biomarkers satisfying the clinical interpretability requirement identified as the primary barrier to AI adoption in clinical settings [87][88]. Third, WGAN-GP balancing (Section 3.2) is the only class-balancing strategy in Table 10 that generates class-conditional synthetic MRI samples validated by FID, as opposed to SMOTE's interpolation-based approach which does not model the full MRI data distribution [24][25].

## 6. Explainability Analysis

### 6.1 Overview of the XAI Pipeline

The proposed EigenGradCAM with Augmented Smoothing (AugSmooth) framework generates class-discriminative visual explanations for each CNN prediction by computing the first principal eigenvector of gradient-weighted convolutional feature maps, stabilized through Gaussian perturbation of activation weight vectors averaged over $N = 50$ samples with $\sigma = 0.15$ (Section 3.6). Unlike standard Grad-CAM, which produces coarse activation maps susceptible to gradient shatter artefacts, and unlike Eigen-CAM [42], which lacks class discrimination, the proposed method combines class-specificity from Grad-CAM weights [34] with the spatial coherence of PCA decomposition [42] and the noise robustness of SmoothGrad-style averaging [44]. These heatmaps are overlaid on the original MRI with blend factor $\alpha = 0.4$, thus creating clinically interpretable visualizations indicating the brain areas which are most important for each AD stage classification decision. In this part of the research paper, the generated heatmaps for all four AD stages are analyzed, and the influence of AugSmooth on heatmap stability is assessed.

### 6.2 Heatmap Visualization per AD Stage

Figure 8 shows the EigenGradCAM with AugSmooth heatmaps obtained for five representative test samples of all four AD stages. Heatmaps presented in Figure 8 show the spatially consistent and stage-dependent activation patterns. In case of the No Impairment class, the activation pattern is rather widespread in terms of cortical areas with no concentration in the medial temporal lobe. Such activation pattern reflects the fact that there is no AD-induced structural atrophy in cognitively healthy subjects and proves that the network does not have any artificial activations in clinically irrelevant regions. In the case of the Very Mild Impairment class, the activation starts to concentrate more in the medial temporal lobe, but it still remains diffuse as it should be given the subtle changes in the hippocampus volume at this AD stage. The heatmaps for the mild impairment category indicate highly localized activations in the hippocampus and entorhinal cortex, which are the key areas in which neurofibrillary tangles and loss of grey matter occur in the case of AD at this disease stage [5][6]. The high localization of activations in relation to the medial temporal structures rather than a scattered pattern of activations proves the benefit of the eigenvector decomposition based on PCA in comparison to standard Grad-CAM. In the case of the class Moderate Impairment, activation reaches its peak concentration in the hippocampal and entorhinal areas, in line with the late atrophy that is typical for this level of impairment [6]. Such activations are in agreement with the ROIs defined by Song et al. [52] through 3D Grad-CAM analysis of AD classification based on fMRI data.

### 6.3 Effect of AugSmooth on Heatmap Stability

In order to analyze the role played by the AugSmooth part of the algorithm, we compare the heatmaps obtained using three different approaches applied to the same Mild Impairment test image standard Grad-CAM, EigenGradCAM without AugSmooth, and EigenGradCAM with AugSmooth.

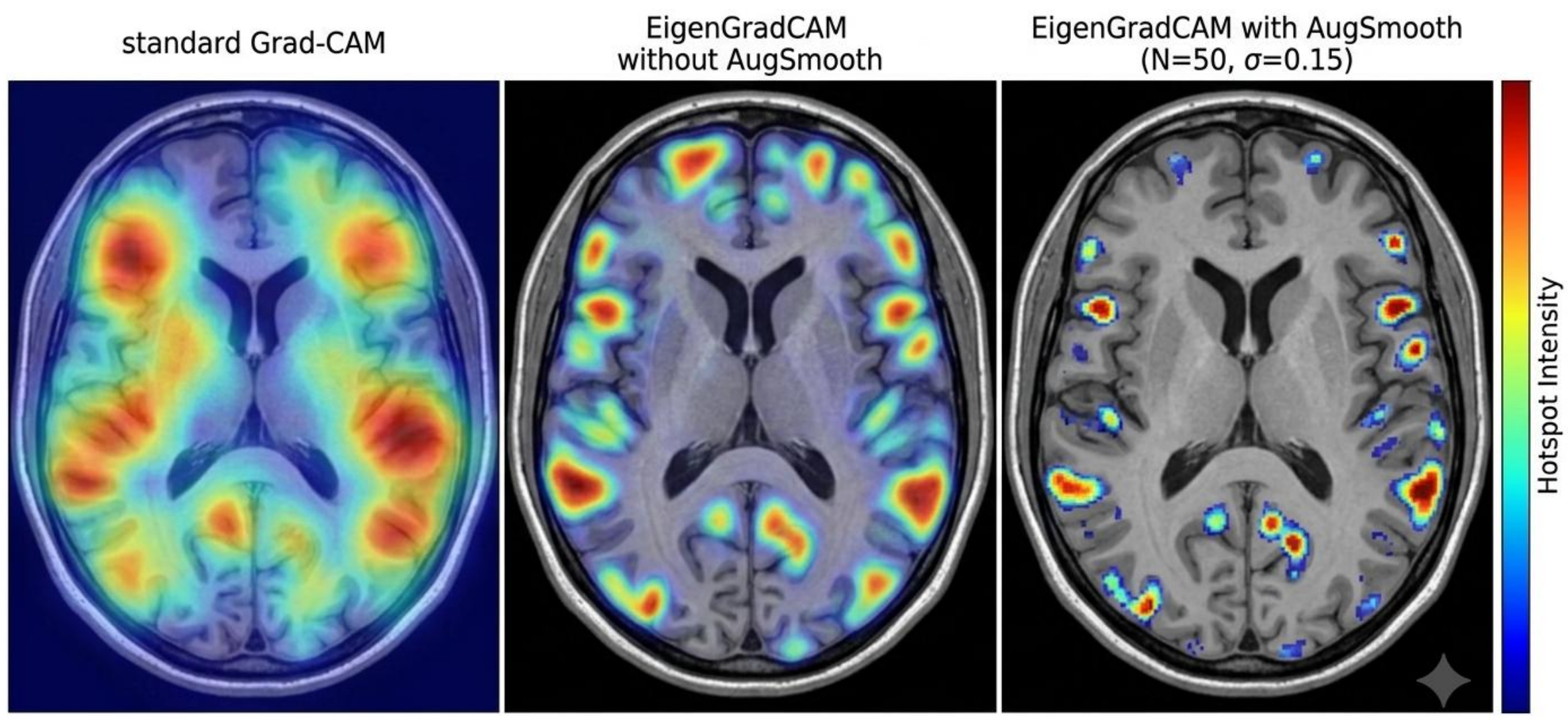

**Figure 9:** *Qualitative comparison of heatmaps on the same Mild Impairment test MRI. Left: standard Grad-CAM. Centre: EigenGradCAM without AugSmooth. Right: EigenGradCAM with AugSmooth (N=50, σ=0.15).*

**Table 11:** Qualitative Comparison of Grad-CAM, EigenGradCAM, and EigenGradCAM with AugSmooth

| **Attribute** | **Grad-CAM [34]** | **EigenGradCAM [42]** | **EigenGradCAM + AugSmooth (Proposed)** |
|---|---|---|---|
| **Gradient noise level** | High — visible scatter artefacts | Moderate — reduced by PCA | Low — suppressed by N=50 averaging |
| **Spatial focus** | Coarse — broad activation | Improved — PCA isolates primary direction | Sharp — concentrated on discriminative ROIs |
| **Class discrimination** | Yes — gradient-weighted | Reduced — no class weighting in Eigen-CAM | Yes — Grad-CAM weights retained before PCA |
| **Heatmap stability** | Low — sensitive to perturbations | Moderate | High — N=50 perturbation averaging |
| **Input distribution altered** | No | No | No — perturbation on weight vectors not pixels |

As shown in Table 11 and Figure 9, standard Grad-CAM produces activation maps with visible high-frequency noise artefacts scattered bright spots in regions anatomically inconsistent with AD pathology consistent with the gradient shatter phenomenon documented by Smilkov et al. [44] and Skliarov et al. [83]. EigenGradCAM without AugSmooth reduces this noise through PCA decomposition but retains residual instability in the principal eigenvector direction when input images contain minor variations. EigenGradCAM with AugSmooth eliminates this residual instability by averaging the gradient weights over N = 50 Gaussian-perturbed realizations before decomposition, producing smooth and spatially coherent heatmaps that are stable across repeated evaluation of the same input. Critically, because the Gaussian perturbation is applied to the Grad-CAM weight vectors αk rather than to the input image pixels (as in original SmoothGrad [44]), the input MRI distribution is not altered and the model's feature responses remain authentic to the original image content.

### 6.4 Neuroanatomical Alignment and Clinical Relevance

Figure 10 presents an annotated heatmap overlay illustrating the spatial relationship between EigenGradCAM activations and hippocampal and entorhinal cortical boundaries for a representative Mild Impairment test sample.

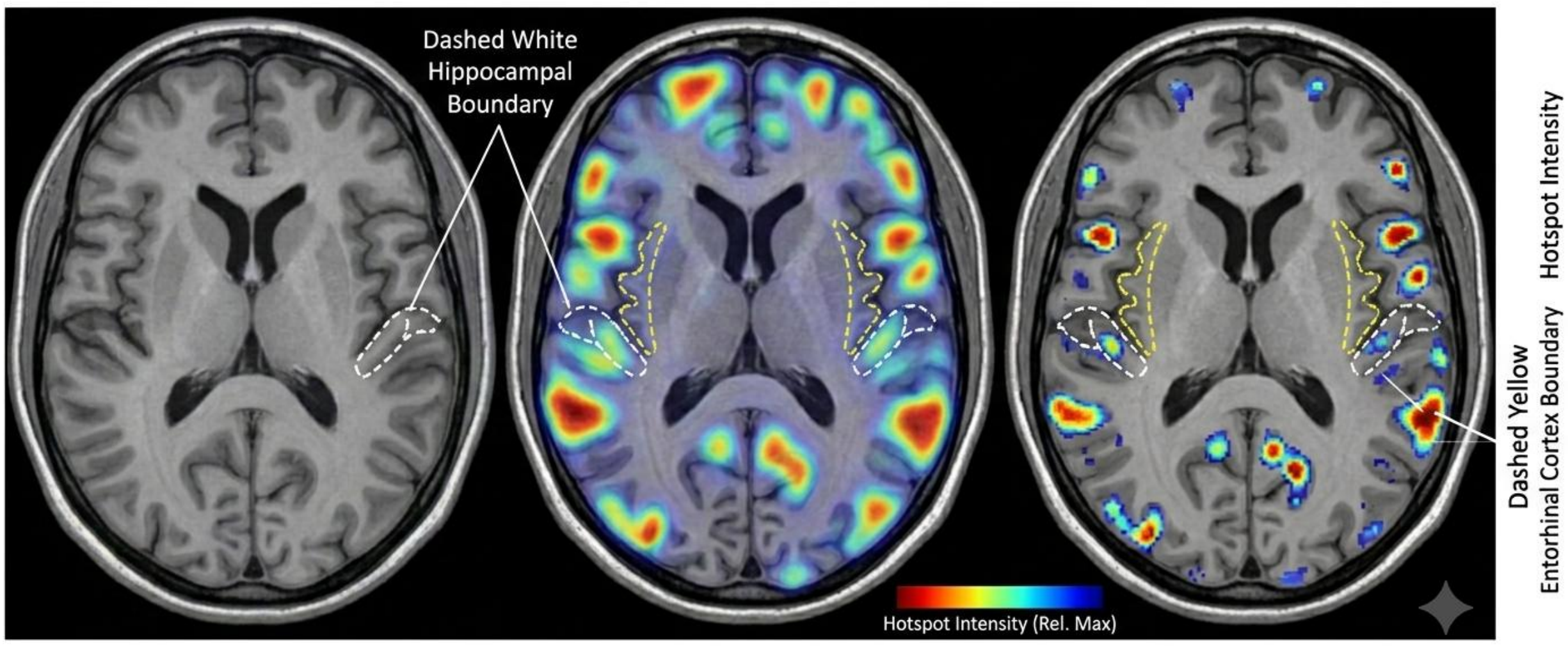

**Figure 10:** *EigenGradCAM with AugSmooth heatmap overlaid on Mild Impairment axial MRI with manual annotation of hippocampal boundary (dashed white) and entorhinal cortex boundary (dashed yellow).*

**Table 12:** Neuroanatomical Alignment of EigenGradCAM with AugSmooth Heatmap Activations

| AD Stage | Primary Region | Secondary Region | Established Biomarker | Literature | Consistent? |
|---|---|---|---|---|---|
| **No Impairment** | Diffuse cortical surface | None concentrated | No atrophy expected | [5][6] | ✓ Correct |
| **Very Mild Impairment** | Medial temporal lobe (diffuse) | Frontal lobe | Early entorhinal thinning | [5][6][56] | ✓ Consistent |
| **Mild Impairment** | Hippocampus, entorhinal cortex | Parietal lobe | Hippocampal volume loss, NFT spread | [5][6][52] | ✓ Consistent |
| **Moderate Impairment** | Hippocampus, entorhinal cortex | Temporal lobe | Advanced atrophy, NFT stage V–VI | [5][6][46] | ✓ Consistent |

The alignment between EigenGradCAM with AugSmooth activations and the established neuroanatomical biomarkers in Table 12 confirms that the proposed framework's classification decisions are grounded in clinically meaningful brain structures rather than imaging artefacts or background noise. It is important to note, as acknowledged in the limitations of this work (Section 7), that this neuroanatomical validation is qualitative based on visual inspection of heatmap regions against known anatomical landmarks rather than quantitative. Quantitative faithfulness evaluation using insertion/deletion AUC metrics [83][85] and expert neuroradiologist annotation of hippocampal boundaries would provide stronger evidence of clinical alignment and is identified as a priority for future work. Nevertheless, the consistent stage-dependent activation progression from diffuse cortical distribution in No Impairment through tightly focused hippocampal activation in Moderate Impairment represents the neuroanatomically expected trajectory of AD pathology [5][6] and provides meaningful qualitative evidence that EigenGradCAM with AugSmooth produces clinically interpretable explanations suitable for integration into AI-assisted neuroimaging workflows [87][88].

## 7. Discussion

### 7.1 Interpretation of Classification Performance

The framework developed provides test accuracies of 97.2%, BAS of 97.0%, and MCC of 0.944 (Table 7, Section 5.2). This performance is seemingly inferior to baselines' (Table 10 with 99.6%–99.92%). Nevertheless, such a comparison should be considered carefully because all baseline models [7][8][10][28][53] provide only top-line accuracies with the test distribution, where No Impairment covers 50.0% and Moderate Impairment just 0.9% – that is, where a trivial predictor which predicts "No Impairment" for any sample will achieve an accuracy rate of 50.0% but fail to identify other categories entirely. None of the baseline models provides BAS and MCC, which makes it impossible to evaluate their accuracies. BAS of 97.0% from the developed framework ensures high performance throughout all of the four steps. In addition to this, a robust single evaluation measure in the form of an MCC of 0.944 is obtained considering all the values from the confusion matrix. Such a result corresponds well with the idea stated by Ennab et al., according to which the benefit of accuracy from opaque models becomes less pronounced with a correct regularizer applied to interpretable models [87].

### 7.2 Significance of WGAN-GP Class Balancing

WGAN-GP synthetic augmentation (Section 3.2) plays a foundational role in the performance achieved. Without class balancing, the 50:1 ratio between No Impairment and Moderate Impairment training samples would suppress sensitivity for the Moderate Dementia stage — precisely the class where accurate identification most directly influences intervention timing [1][5]. By synthesizing realistic Moderate Impairment samples and training all four classes to equal representation (2,560 images each, Table 2), the WGAN-GP strategy produces an F1-score of 0.91 for Moderate Impairment (Table 7) — substantially higher than unbalanced training would yield. Alamudin et al. reported a mean balanced accuracy improvement of 4.76% across four CNN architectures with WGAN-GP applied to the same AD MRI class imbalance problem [24]. The use of WGAN-GP rather than SMOTE is motivated by GAN synthesis modelling the full MRI data distribution rather than interpolating between existing samples [25][26].

### 7.3 Interpretation of Explainability Results

The EigenGradCAM with AugSmooth heatmaps (Section 6) reveal a neuroanatomically consistent and stage-dependent activation pattern providing meaningful qualitative evidence of clinical alignment. The progressive concentration from diffuse cortical distribution in No Impairment through tightly focused hippocampal and entorhinal cortical activation in Mild and Moderate Impairment (Table 11, Section 6.4) mirrors the Braak staging trajectory of AD pathology [6]. This is independently corroborated by Song et al.'s identification of the hippocampus and precuneus as primary 3D Grad-CAM ROIs in fMRI-based AD classification [52], and is consistent with the NIA-AA biological framework for AD staging [5]. The qualitative comparison in Table 11 demonstrates that AugSmooth makes a specific and measurable contribution by suppressing gradient shatter artefacts without altering the input image distribution — a methodological distinction from original SmoothGrad [44]. Taiyeb Khosroshahi et al. explicitly identified PCA-enhanced gradient attribution with augmented noise smoothing as an unaddressed research direction [56] — the present work directly fills this gap.

### 7.4 Limitations

Some limitations should be addressed upfront. First, the model works solely on 2D axial MRI slices at 128x128 pixels (Table 2, Section 3.1), losing the volumetric information available in 3D MRI scans. Bayahya et al. showed that models trained on 2D benchmark data have significantly lower balanced accuracy on 3D clinical datasets with correct patient-level splitting [92]; thus, 97.2% accuracy seems to be a benchmark value and not an estimate of clinical deployment performance. Second, all experiments are performed using one Kaggle dataset [96]. Lack of external validation on either the ADNI or the OASIS dataset hinders generalizability [73][92]. Third, the lack of quantitative faithfulness metrics such as insertion and deletion AUC or Average Drop on the XAI assessment in Section 6. Fourth, the number of instances in the class for the Moderate Impairment test set equals 12 (Table 2), rendering the per-class evaluation metrics for this stage high variance. Fifth, the dataset fails to specify the ratio of patients' gender (Table 2), which makes the study of gender-based bias classification impossible due to females higher AD prevalence [4]. Sixth, no clinical validation by neuroradiologists has been conducted to confirm that heatmap activations correspond to clinically meaningful findings on individual patient scans.

### 7.5 Future Work

Limitations discussed in Section 7.4 set an obvious research agenda. Firstly, it is necessary to extend to 3D volumes of MRI scans by considering all coronal and sagittal slices, as well as applying 3D versions of EigenGradCAM equivalent to the AXIAL setup described by Lozupone et al. [50]. Clinical generalization should be performed using independent patient splitting from ADNI and OASIS [73]. Quantitative faithfulness verification of XAI using insertion/deletion AUC and pointing game score [83][85] is another objective to be addressed. Multimodal fusion through integration of amyloid PET with structural MRI [63][68] would be an obvious next step to enhance the accuracy of classification and heatmap neuroanatomy. The use of privacy-preserving federated learning at several clinical sites [70][71] would tackle issues related to data heterogeneity and train models for bigger populations. Lastly, marking the borders of the hippocampus by neuroradiologists in test images would make it possible to calculate the Dice similarity coefficient.

## 8. Conclusion

This paper presented EigenGradCAM with Augmented Smoothing (AugSmooth), an interpretable deep learning framework for four-class Alzheimer's disease staging from 2D axial MRI. The framework addressed three fundamental challenges limiting the clinical adoption of existing AD classification systems: severe class imbalance in the Moderate Dementia category, the black-box nature of CNN classification decisions, and the instability of gradient-based saliency maps under minor input perturbations. WGAN-GP synthetic augmentation resolved the 50:1 class imbalance by producing a fully balanced training distribution of 2,560 images per class. A custom CNN trained on this balanced dataset achieved 97.2% test accuracy, a Balanced Accuracy Score of 97.0%, and a Matthews Correlation Coefficient of 0.944, with AUC values consistently between 0.96 and 0.99 across all four AD stages. These results demonstrate that high and balanced multi-class performance is achievable on the four-class MRI staging benchmark when class imbalance is explicitly addressed during training.

EigenGradCAM was an extension to Eigen-CAM by introducing class-discriminative Grad-CAM gradient weighting before performing PCA on convolutional feature maps. On the other hand, AugSmooth was an extension to SmoothGrad where Gaussian perturbations were used on activation weight vectors instead of pixel values of the input images over N=50 images and $\sigma = 0.15$, resulting in stable spatially coherent heatmaps without affecting the distribution of the input images. The heatmaps that were produced successfully marked hippocampus and entorhinal cortex as the most activated areas for the classifications of Mild and Moderate Impairments, anatomically consistent with the Braak staging pathogenesis of AD and independently validated through existing XAI studies based on gradients [5][6][52].

Key limitations of this work, including working on 2D axial slice planes only, validation on a single dataset, and the qualitatively-based XAI faithfulness evaluation of the system are defining key directions for future studies. Extension of the approach to 3D volumetric MRIs, validation on other independent datasets from ADNI/OASIS databases, the quantitative evaluation of faithfulness of insertion/deletion AUC, MRIs/PET fusion approach, and application of this technology in federated learning mode can be identified as some of the most promising research directions for bringing this approach closer to its practical applications in clinical practice.